\documentclass[conference]{IEEEtran}
\IEEEoverridecommandlockouts

\usepackage{cite}
\usepackage{amsmath,amssymb,amsfonts}
\usepackage{graphicx}
\usepackage{textcomp}
\usepackage[table,xcdraw,dvipsnames]{xcolor}
\def\BibTeX{{\rm B\kern-.05em{\sc i\kern-.025em b}\kern-.08em
    T\kern-.1667em\lower.7ex\hbox{E}\kern-.125emX}}

\usepackage{soul}
\usepackage{xspace}
\usepackage{subcaption}
\usepackage{algorithm}
\usepackage{algpseudocode}
\usepackage{tikz}
\usepackage{array}
\usepackage{enumitem}
\usepackage{multirow}
\usepackage{setspace}
\usepackage[bookmarks=true,breaklinks=true,letterpaper=true]{hyperref}

\newcommand*\circled[1]{\tikz[baseline=(char.base)]{
        \node[shape=circle,draw,inner sep=0.1pt] (char) {#1};}}

\newcommand{\fig}[1]{Figure~\ref{#1}}
\newcommand{\equat}[1]{Equation~\ref{#1}}
\newcommand{\sect}[1]{Section~\ref{#1}}
\newcommand{\tab}[1]{Table~\ref{#1}}
\newcommand{\algo}[1]{Algorithm~\ref{#1}}

\begin{document}

\newcommand{\proposed}[0]{vTrain\xspace}
\title{
    \huge vTrain: A Simulation Framework for Evaluating Cost-effective and Compute-optimal Large Language Model Training
    }

\author{
\IEEEauthorblockN{Jehyeon Bang\IEEEauthorrefmark{2}\quad
                Yujeong Choi\IEEEauthorrefmark{2}\quad
                Myeongwoo Kim\IEEEauthorrefmark{1}\quad
                Yongdeok Kim\IEEEauthorrefmark{1}\quad
                Minsoo Rhu\IEEEauthorrefmark{2}\quad
    }
\\
\IEEEauthorblockA{\IEEEauthorrefmark{2}KAIST\\School of Electrical Engineering
\\\texttt{\{jehyeon.bang, yjchoi0606, mrhu\}@kaist.ac.kr}}
\\
\IEEEauthorblockA{\IEEEauthorrefmark{1}Samsung Advanced Institute of Technology
\\\texttt{\{k.myeong-woo, yd.mlg.kim\}@samsung.com}}
}

\maketitle

\begin{abstract}

As large language models (LLMs) become widespread in various application domains, a critical challenge the AI community is facing is how to train these large AI models in a cost-effective manner. Existing LLM training plans typically employ a heuristic based parallel training strategy which is based on empirical observations rather than grounded upon a thorough examination of the search space of LLM parallelization. Such limitation renders existing systems to leave significant performance left on the table,  wasting millions of dollars worth of training cost.
This paper presents our profiling-driven simulator called \proposed, providing AI practitioners a fast yet accurate software framework to determine an efficient and cost-effective LLM training system configuration. We demonstrate \proposed's practicality through several case studies, 
e.g., effectively evaluating optimal training parallelization strategies that balances training time and its associated training cost, 
efficient multi-tenant GPU cluster schedulers targeting multiple LLM training jobs,
and determining a compute-optimal LLM model architecture given a fixed compute budget. 
    
    \end{abstract}
\section{Introduction} 
\label{sect:intro}

Transformer-based~\cite{transformer} large language models (LLMs) like ChatGPT~\cite{chatgpt}, Gemini~\cite{gemini}, LLaMA~\cite{llama2}, and PaLM~\cite{palm} are taking the world by storm. Driven by the rapidly increasing model parameter size as well as training token size, 
these large autoregressive AI models have demonstrated phenomenal performance on many tasks using a variety of
evaluation protocols such as zero-shot, few-shot, and fine-tuning~\cite{gpt3}.

Unfortunately, a critical challenge that the AI community is facing is how to train these LLMs in a cost-effective  manner. 
State-of-the-art LLMs already reached hundreds of billions of parameters requiring trillions of tokens for training~\cite{chinchilla}. Such massive compute requirements for training LLMs rendered the latest AI training cluster to be provisioned with ExaFLOPS scale computational throughput using tens of thousands of GPUs~\cite{rsc,azure_supercom}. 
Because of the enormous compute budget required for training LLMs, it is practically only possible to train these models ``once'' (e.g., training LLMs like ChatGPT incurs tens of millions of dollars~\cite{mt_nlg, gpt3_training_time}). Consequently, accurately determining the hyperparameters of both the AI training system as well as the LLM model architecture (e.g., number of GPUs, intra-/inter-node parallelization and communication mechanisms, LLM model parameter size, and total number of tokens to train) becomes  critical to optimize not only the LLM training time but more crucially the monetary training ``cost'' associated with it.

\begin{figure}[t!] \centering
    \includegraphics[width=0.39\textwidth] {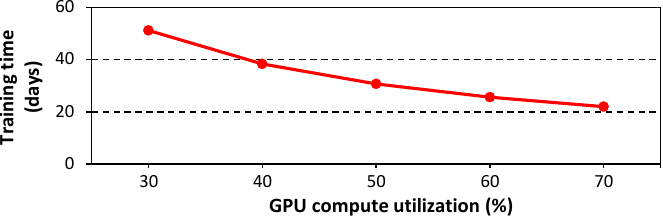}
    \vspace{0.2em}
    \caption{Wall clock training time of GPT-3 (175B parameters) as a function of GPU compute utilization, assuming 1,024 NVIDIA A100 GPUs are used for training. 
    GPU compute utilization refers to the achieved FLOPS relative to the maximum FLOPS.
    Training time is primarily determined by dividing up ``the total number of FLOPs to train an LLM'' with ``the aggregate, effective FLOPS available for training across the 1,024 A100 GPUs''. We estimate training time by changing the effective FLOPS of A100 GPUs and derive the training cost based on AWS EC2 P4d GPU instance pricing information~\cite{amazon_p4d}.}
    \label{fig:compute_util_to_training_time}
    \vspace{-1.3em}
\end{figure}

Given this landscape, a rich set of prior literature explored various LLM parallelization and communication mechanism specifically targeting transformer-based language models~\cite{mt_nlg, megatron_lm, megatron2, gshard}. A popular way of parallelizing autoregressive LLMs today is to employ tensor parallelism~\cite{megatron_lm} for \emph{intra}-node training while combining data parallelism and pipeline parallelism~\cite{gpipe, pipedream} for \emph{inter}-node training (\sect{sect-2B}). Such heuristic based training parallelization strategies have so far proven successful in training conventional LLMs with reasonable efficiency. Nonetheless, the principles behind these heuristic based parallelization are typically founded upon empirical observations or intuitive design choices rather than being grounded upon a rigorous evaluation of the design space of LLM parallelization. This leads to suboptimal GPU utilization which leaves significant performance left on the table (\sect{sect:cost-optimal-case-study}). In \fig{fig:compute_util_to_training_time}, we show the change in LLM training time as a function of GPU's compute utilization when training GPT-3 (175B) using 1,024 A100 GPUs. Notice how the training time is increased by $8$ days when the average GPU compute utilization is merely degraded by $10\%$ (from 50\% to 40\%), resulting in an additional training cost of millions of dollars. 
Unfortunately, 
finding the most optimal and cost-effective hyperparameters of the training system configuration by exhaustively searching through its design space is extremely challenging because of the scale in which these LLMs must be trained. Such limitations render AI practitioners to employ the previously validated, known-good, yet sub-optimal heuristic based training recipes.

In this work, we present a profiling-driven simulator which guides the evaluation of a cost-effective and compute-optimal LLM training system (aka {\bf V}irtual {\bf Train}, \emph{\proposed}\footnote{\proposed is open-sourced at \href{https://github.com/VIA-Research/vTrain}{https://github.com/VIA-Research/vTrain}.} in short).
As discussed above, exhaustively searching through the enormous design space of LLM's training system hyperparameters over real AI training clusters is impractical given the enormous cost it will incur.
Instead, we propose a profiling-based simulation methodology that accurately estimates the LLM training time of each design point in a high-performance manner, rendering the optimal LLM training system configuration to be determined in several tens of minutes using a high-end multi-core CPU server.
While being extremely fast, the design of \proposed is driven by the following key observations that enable an accurate estimation of LLM's training time.
Latest AI algorithms are expressed as a direct acyclic graph where each graph node represents a neural network layer.
In essence, the order in which the graph node gets executed determines the AI model's execution dataflow.
For LLM inference, it is challenging to accurately estimate the model's execution dataflow because the autoregressive nature of LLMs render the LLM graph nodes' execution order to be determined dynamically at runtime.
In contrast, the execution order of LLM graph nodes for training is \emph{precisely} defined at compile time so it is possible for \proposed to statically determine how many LLM graph nodes to execute and the order in which they are executed.
Additionally, the execution time of each individual LLM graph node (each layer) over a target GPU architecture is highly deterministic and exhibits little variance across different runs.
We employ the aforementioned properties of LLM training to develop a sophisticated profiling-based framework that translates how each individual LLM graph nodes (layers) are decomposed into low-level GPU CUDA kernels, in accordance to the LLM parallelization strategy employed, which we utilize to estimate the execution time of a given iteration of LLM training.
To demonstrate \proposed's practicality, we present several case studies tackling critical research problems of LLM training:

\begin{itemize}
\item ({\bf Cost-effective LLM training plan}) Given a target LLM, training token size, and compute budget (i.e., total number of GPUs), what is the most optimal training parallelization strategy that minimizes wall clock training time and its associated training cost?

\item ({\bf Cost-effective multi-tenant LLM scheduling}) Given multiple LLM training jobs that share a GPU cluster, what is an efficient scheduling algorithm that maximize GPU utilization while minimizing job completion time?

\item ({\bf Compute-optimal LLM model design}) What is the largest LLM one can train that satisfies  the Chinchilla scaling law~\cite{chinchilla} (i.e., an LLM providing the best model accuracy) given a fixed compute  and training time budget? In other words, what is the best LLM one can develop within $N$ days using $M$ GPUs?

\end{itemize}

Overall, \proposed is a fast yet accurate  software framework that guides AI practitioners the development of an efficient and cost-effective LLM training configuration. In the following section, we provide the necessary background on LLM training and discuss key motivations behind our work.

\section{Background}

\subsection{Transformer-based Large Language Models (LLMs)}
\label{sect2A}

\begin{figure}[t] \centering
    \includegraphics[width=0.42\textwidth]{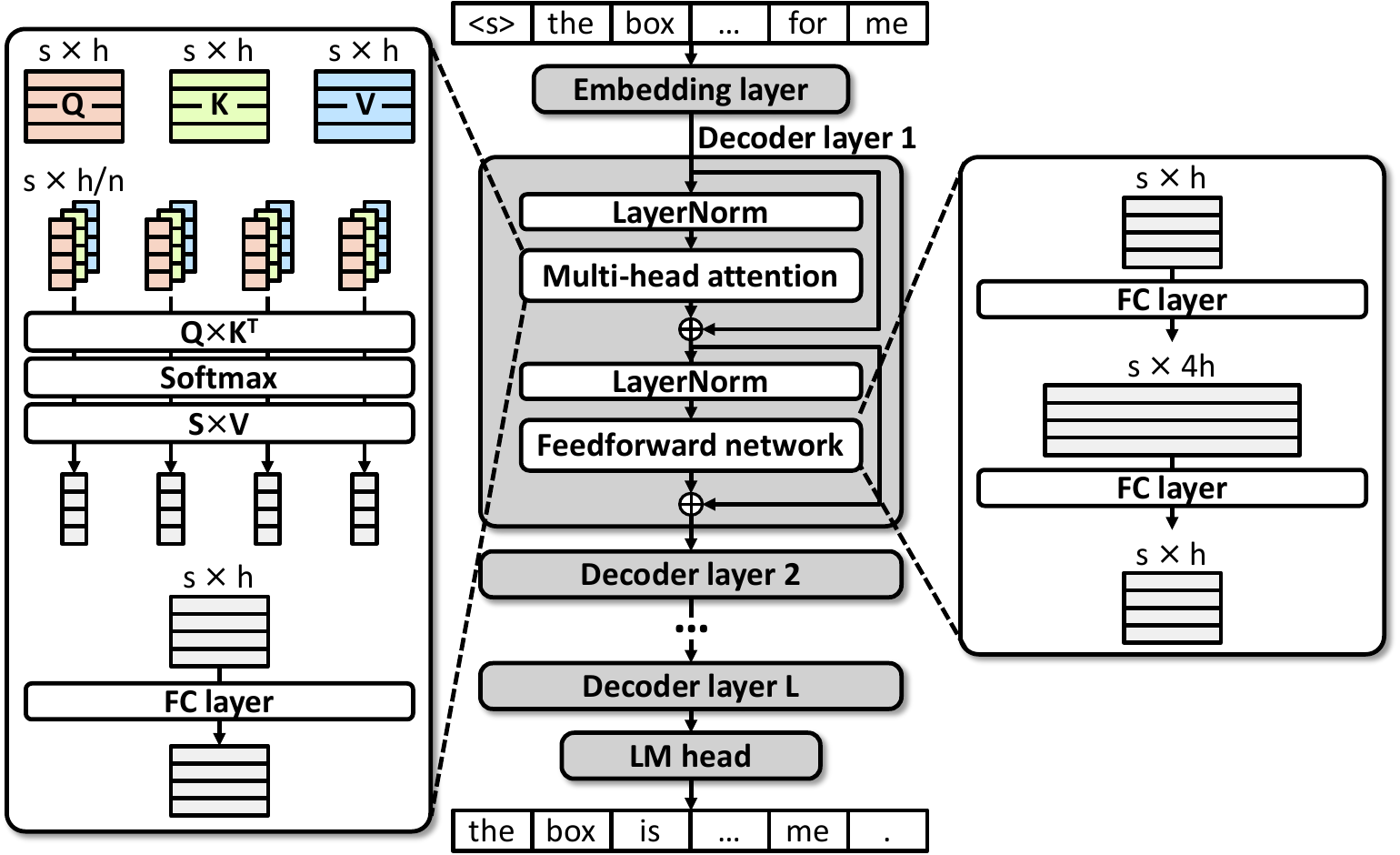}
    \vspace{0.2em}
    \caption{A transformer-based, decoder-only LLM architecture.}
    \label{fig:llm_architecture}
        \vspace{-0.5em}
\end{figure}

\textbf{LLM model architecture.} Practically all recent LLMs are based on transformers~\cite{transformer}, employing a decoder-only architecture~\cite{gpt}.
A decoder-only LLM consists of an embedding layer, several stacks of decoder layers, and a language modeling (LM) head,
    which is characterized by the following hyperparameters (\fig{fig:llm_architecture}): 
    hidden size ($h$),  number of layers ($L$),  maximum sequence length ($s$), and number of attention heads ($n$).
The embedding layer consists of word embeddings and positional embeddings, which are conceptually a look-up table indexed 
    by the input token's ID and its position within the input sequence, respectively. 
The $s$ input tokens are converted into embedding vectors by adding the two types of embeddings at the end of the embedding layer, 
    resulting in an embedding matrix with a size of ($s \times h$).

The decoder layer includes two major blocks: multi-head attention (MHA) and feedforward network (FFN).
MHA block first generates \textit{Query} (Q), \textit{Key} (K), and \textit{Value} (V) matrices, each with a size of ($s \times h$).
The QKV matrices are then divided into $n$ attention heads along the $h$ dimension, and the attention scores are derived for each attention head in parallel.
The results of attention heads goes through a FC (fully connected) layer and subsequently fed into the following FFN block after normalization.
The FFN block is composed of two consecutive FC layers with an intermediate activation size of $4h$.
An LLM stacks $L$ decoder layers in sequence, and the output of the last decoder layer is fed into the LM head.
Finally, the next token for each position is processed in the LM head by multiplying the decoder layer's output with the transpose of the word embedding weight matrix. 

\textbf{Compute-optimal LLM training (aka Chinchilla scaling law).} Although recent LLMs have grown in its parameter size rapidly 
    from $1.5$B (GPT-2~\cite{gpt2}) in $2019$ to $540$B (PaLM~\cite{palm}) in $2022$,
    the total number of tokens used for training has remained almost constant. 
For example, Megatron-Turing NLG (MT-NLG)~\cite{mt_nlg} contains 530B parameters, three times the size of the $175$B parameter GPT-3~\cite{gpt3},
    but both models are trained with approximately $270$B tokens~\cite{mt_nlg, gpt3}.
Interestingly, recent work by Hoffmann et al.~\cite{chinchilla} observed that \emph{under}-training an LLM cannot fully reap out the algorithmic potential of the LLM,
    suggesting that the model size and the number of training tokens should be scaled equally.
Such ``compute-optimal'' training, also known as the \emph{Chinchilla scaling law}, highlights the existence of a power law relationship between the model size,
    the number of training tokens, and the available compute budget (detailed further in \sect{sect:chinchilla}).
According to this scaling law, it is possible to identify the optimal model and training dataset size that achieves the minimum training loss within a given compute budget (the ``Chinchilla point'').

\subsection{LLM Parallelization Strategies} \label{sect-2B}

\begin{figure}[t!]
    \centerline{\includegraphics[width=\columnwidth]{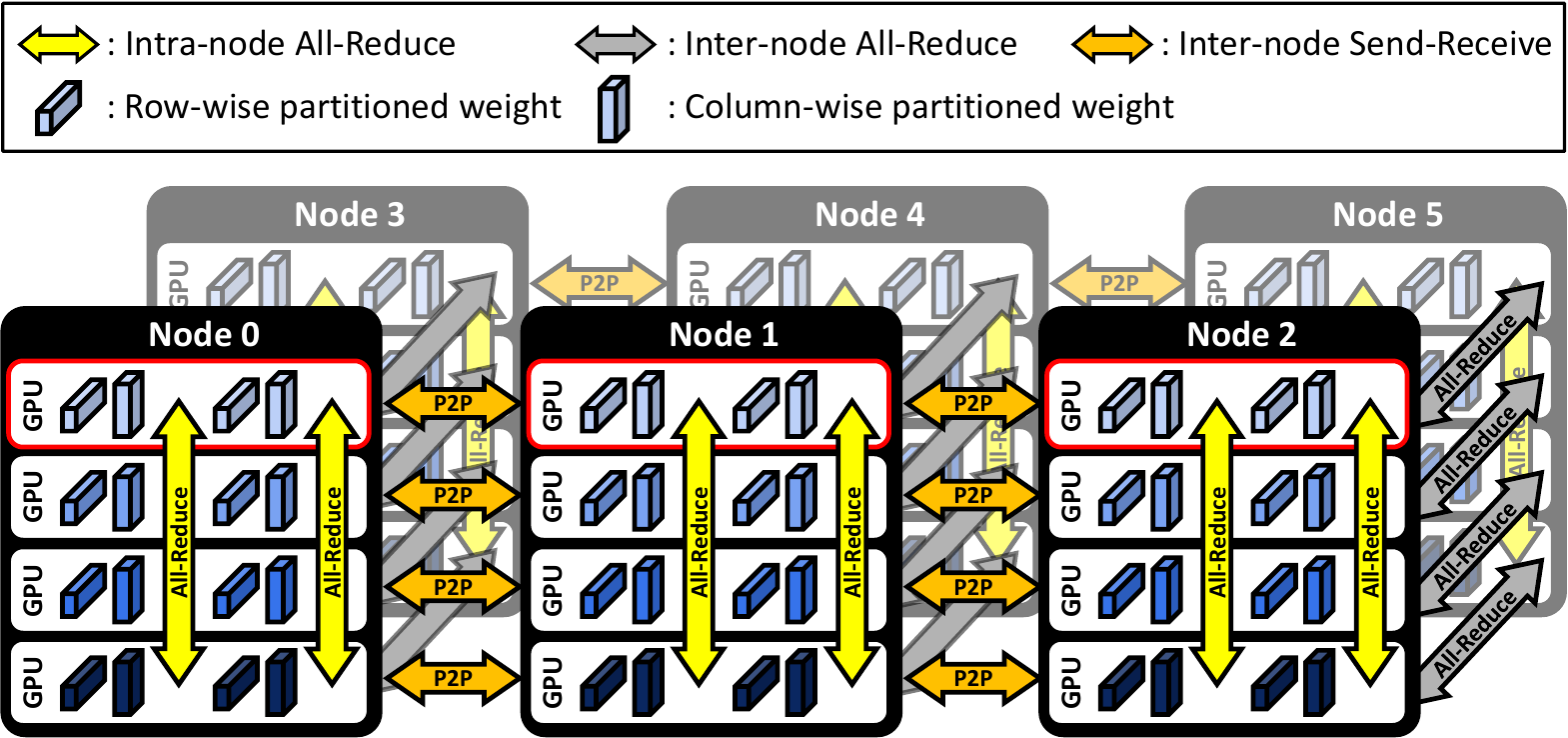}}
    \vspace{0.4em}
    \caption{
        An LLM training system employing 3D parallelism.
        Example combines 4-way tensor parallelism (\emph{intra}-node $4$ GPUs invoking the yellow colored All-Reduce), 
            2-way data parallelism (the node pairs invoking the gray colored \emph{inter}-node All-Reduce),
            and 3-way pipeline parallelism (the three nodes [0,1,2] and [3,4,5] invoking the orange colored \emph{inter}-node Send-Receive).
        In the rest of this paper, a (\emph{t}, \emph{d}, \emph{p})-way 3D parallelism refers to a training system configuration 
            employing \emph{t}-way tensor, \emph{d}-way data, and \emph{p}-way pipeline parallelism, i.e., example illustrates (4,2,3)-way 3D parallelism.
    }
    \label{fig:3d_parallelism}
        \vspace{-0.8em}
\end{figure}

\textbf{Data parallelism.} Under data parallelism, the input training dataset is partitioned across all the workers but each worker has the same model replica~\cite{li13pytorch}.
As such, each worker performs the forward and backward pass of training independently of the other workers.
Before updating the model weights, however, all the workers need to synchronize the weight gradients with each other using a \emph{gradient reduction}
    (implemented using an \emph{All-Reduce} operator~\cite{allreduce}, the gray colored All-Reduce operations in \fig{fig:3d_parallelism}) 
    to ensure that all workers have an identical version of the model weights. 

\textbf{Tensor parallelism.} Unlike data parallelism, all workers in model parallel training work on an identical batch of the input training dataset 
    but each worker is allocated with a different slice of the model parameters.
Tensor parallelism~\cite{megatron_lm} is a specific form of model parallelism tailored for LLMs, 
    which splits a FFN block's first and second FC weight matrices along its columns and rows, respectively
    (the row-wise/column-wise partitioned weights in \fig{fig:3d_parallelism}).
The benefits of such parallelization strategy is that it reduces the inter-layer synchronization overhead to a single All-Reduce operation
    during forward and another All-Reduce during the backward pass.
The MHA block can similarly employ tensor parallelism and incur a single All-Reduce operation during forward and backward pass.
It is worth pointing out that the All-Reduce operation after each MHA block and FFN block creates a sequential dependency 
    between its respective computation and communication operation 
    (the two yellow colored, intra-node All-Reduce operations within a given node in \fig{fig:3d_parallelism}).

\textbf{Pipeline parallelism.} Pipeline parallelism is another form of model parallelism.
Unlike tensor parallelism, pipeline parallelism partitions LLM's model parameters at the granularity of individual layers~\cite{gpipe,pipedream}.
Each worker is assigned with a mutually exclusive set of consecutive layers of the LLM and is responsible for performing
    the forward pass of its assigned layers, sending the final output activation values to the next adjacent worker. 
During the backward pass, each worker sends/receives input gradients to/from its adjacent workers 
    but in the opposite direction of the forward pass (orange colored inter-node Send-Receive operations in \fig{fig:3d_parallelism}).

\begin{figure*}[t!] \centering
    \centerline{\includegraphics[width=0.78\textwidth]{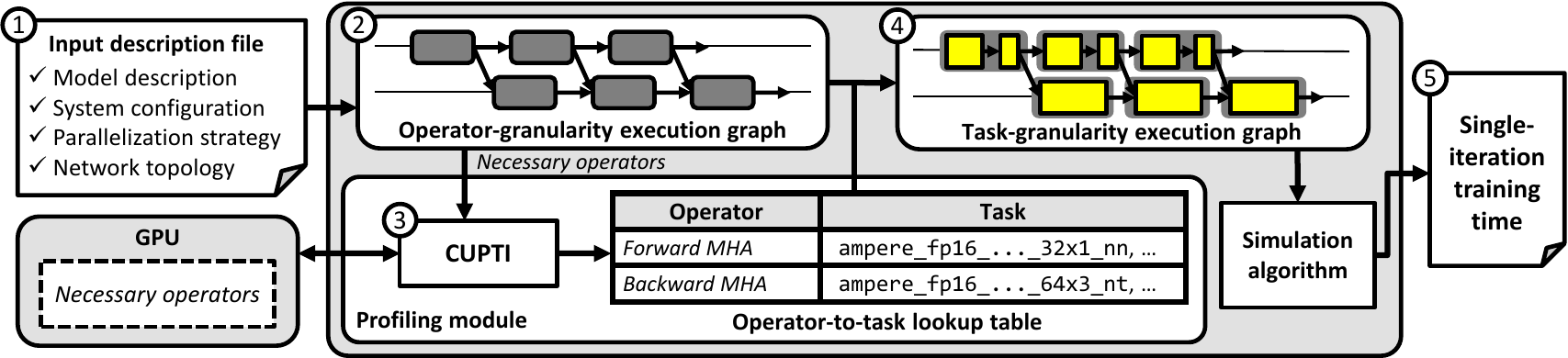}}
    \vspace{0.1em}  
    \caption{Key components of \proposed and its simulation flow.}
    \label{fig:framework_overview}
        \vspace{-0.8em}   
\end{figure*}

GPipe~\cite{gpipe} proposes pipeline parallel training which splits an input batch into smaller \emph{micro}-batches, 
    allowing the training iteration of different micro-batches to be pipelined.
The waiting time for other workers (the pipeline \emph{bubbles}) can be reduced 
    by increasing the number of micro-batches but at the cost of increased memory footprint.
To alleviate such overhead, PipeDream~\cite{pipedream,pipedream_bw} proposed an alternative schedule 
    called one-forward-one-backward (1F1B), where the memory footprint of each worker is reduced 
    by limiting the number of in-flight micro-batches to the pipeline depth 
    (we revisit GPipe/1F1B later in \fig{fig:pipeline_schedule}). 
In general, the latency incurred in pipeline parallelism's inter-stage communication is negligible.
This is because the Send-Receive communication simply exchanges data across two workers and 
    is less sensitive to the interconnection bandwidth.
Rather, overall performance is governed by the amount of pipeline bubbles and 
    more sensitive to the micro-batch size and the pipeline scheduling algorithm.

\textbf{3D parallelism.} State-of-the-art LLMs suffer from a memory capacity bottleneck
    where the latest model's parameter size greatly exceeds the GPU memory size~\cite{deepspeed, mt_nlg}.
As such, a hybrid of data/tensor/pipeline parallelism, commonly referred to as \emph{3D parallelism} (\fig{fig:3d_parallelism}),
    has been widely adopted in recent literature on LLM training~\cite{gshard,gpt3,megatron_lm,mt_nlg}.
Among the three parallelization strategies, tensor parallelism generally invokes the largest communication size 
    while data and pipeline parallelism incurs relatively less communication overhead.
Consequently, state-of-the-art LLM training systems like Megatron-LM~\cite{megatron_lm}, MT-NLG~\cite{mt_nlg}, and OPT-175B~\cite{opt} 
    generally employ tensor parallelism for intra-node training because high-end GPU server nodes from NVIDIA are provisioned 
    with high bandwidth communication links like NVLink~\cite{nvlink}.
For inter-node parallel training, standard practice is to employ a mix of data and pipeline parallelism
    as the workers are provided with relatively less communication bandwidth over Ethernet or InfiniBand.

\section{\proposed: A Profiling-based Software Framework for Simulating LLM Training Time}

\subsection{Simulation Framework Overview}
\label{sect:framework_overview}

\fig{fig:framework_overview} provides a high-level overview of \proposed's simulation flow. An input description file is provided to the simulator which contains the target LLM to evaluate, total number of GPUs available, the training system configuration (e.g., number of GPUs within a node, intra-/inter-node communication bandwidth, $\ldots$), and the LLM parallelization strategy (e.g., (\emph{t}, \emph{d}, \emph{p})-way 3D parallelism) to explore (\circled{1}). 
\proposed then constructs a high-level \emph{operator-granularity execution graph} which represents what are the computation and communication operators to execute, as defined by the LLM model architecture and the parallelization strategy employed (\circled{2}). 
Using the high-level operator-granularity execution graph, \proposed utilizes the profiling module to determine what are the low-level CUDA kernels (tasks) to execute for each of the operators in the operator-granularity execution graph and also measures the wall clock time latency incurred in executing that CUDA kernel, 
constructing a \emph{operator-to-task lookup table} (\circled{3}). Using this lookup table, \proposed translates the high-level operator-granularity execution graph into a low-level \emph{task-granularity} execution graph (\circled{4}) encapsulating which CUDA kernels should be executed for a given operator-granularity execution graph vertex. Finally, \proposed simulates the single-iteration training time by referring to the task-granularity execution graph and the operator-to-task lookup table (\circled{5}).

\subsection{(High-level) Operator-granularity Execution Graph} \label{sect-3B}

The operator-granularity execution graph is designed to represent which operations need to be executed for an LLM and in what order. A graph vertex within the operator-granularity  graph (referred to as a \emph{layer-node}) represents the \emph{operator} to execute for that layer-node and the graph edges that connect layer-nodes determine the dependencies between them, i.e., the execution order. The purpose of constructing the operator-granularity graph is to determine not only the
computation operators (e.g., forward/backward pass of a MHA or FFN layer) but more importantly the communication operators (e.g., All-Reduce, Send-Receive) to execute, which is determined by the 3D parallelism strategy employed. 
As discussed in \sect{sect-2B}, each parallelism strategy introduces a different communication pattern. To accurately simulate an LLM training iteration, \proposed needs to determine  (1) the \emph{type} of communication operators to launch, (2) \emph{when} those communication operators should be executed, and (3) what are the inter-operator dependencies enforced by the insertion of such communication operators. Below we discuss how \proposed augments the operator-granularity graph with the required communication operators per each parallelism type.

\begin{figure}[tb!] \centering
\begin{subfigure}[b]{0.85\columnwidth}
    \includegraphics[width=\columnwidth]{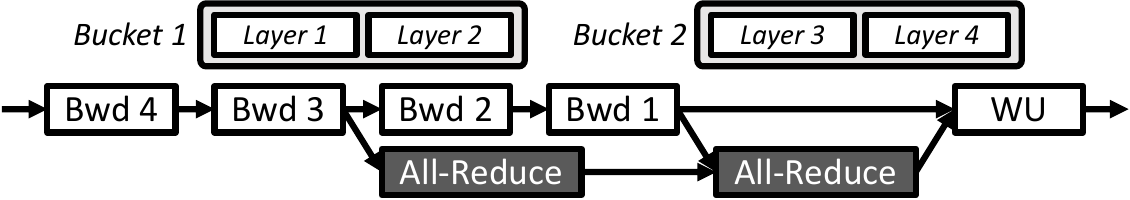}
    \caption{With gradient bucketing}
\end{subfigure}
\begin{subfigure}[b]{0.85\columnwidth}
\vspace{1em}
    \includegraphics[width=\columnwidth]{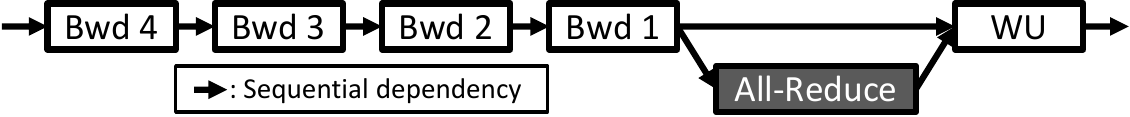}
    \caption{Without gradient bucketing}
\end{subfigure}
\vspace{0em}
\caption{Inserting All-Reduce operators for data parallel training when gradient bucketing is (a) enabled and (b) disabled. ``\texttt{Bwd i}" represents the $i^{th}$ layer's backward pass and ``\texttt{WU}" refers to the weight update pass. Example in (a) assumes that layer ($1\&2$) and ($3\&4$) are grouped into a bucket.}
\label{fig:framework_data_parallel}
\vspace{-1.5em}
\end{figure}

\textbf{Data parallelism.} In data parallel training, an All-Reduce operation is performed targeting the weight gradients. 
State-of-the-art ML frameworks like PyTorch DDP~\cite{li13pytorch} provides an optimization called \emph{gradient bucketing} which helps overlap the All-Reduce communication latency with the computation time of LLM backward pass.
When gradient bucketing is enabled, the weight gradients are separated into several buckets and the software runtime kicks off the All-Reduce communication operator for a gradient bucket whenever any given bucket is ready for an All-Reduce. Such optimization helps  the All-Reduce latency for a target gradient bucket to be hidden inside the other gradient buckets' derivation during backward pass. \proposed properly models such optimization by assigning the proper number of gradient buckets as defined in the LLM model description file, identical to how PyTorch DDP does it. Using such information, an All-Reduce operator for each bucket is properly inserted into the operator-granularity execution graph, as depicted in  \fig{fig:framework_data_parallel}(a). When gradient bucketing is disabled, All-Reduce is required only once at the end of the backward pass and \proposed properly models such behavior as illustrated in \fig{fig:framework_data_parallel}(b).

\begin{figure}[tb!] \centering
    \centerline{\includegraphics[width=0.98\columnwidth]{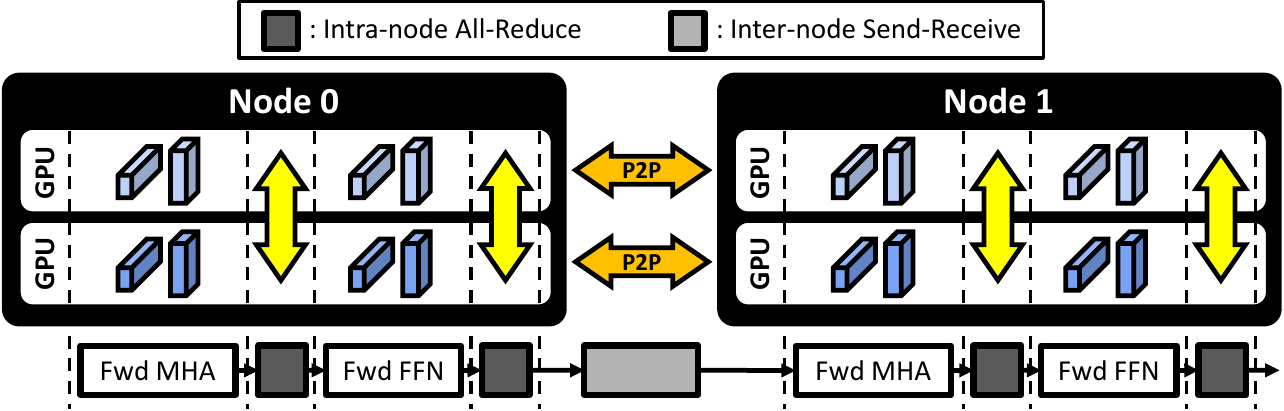}}
    \vspace{0.3em}
    \caption{Example inserting intra-node All-Reduce and inter-node Send-Receive  operators when tensor/pipeline parallelism is employed.}
    \label{fig:tensor_and_pipeline}
 \vspace{-0.3em}   
\end{figure}

\textbf{Tensor parallelism.} As described in \sect{sect-2B}, tensor parallelism alternately splits an LLM decoder layer's weight matrices along the column and row dimension, invoking an All-Reduce operation for every two consecutive matrix multiplication operations of MHA and FFN blocks. \proposed models such communication operation by inserting an All-Reduce operator after the layer-nodes of MHA and FFN blocks for both forward and backward passes (dark gray operator in \fig{fig:tensor_and_pipeline}). These All-Reduce operations are performed in an \emph{intra}-node manner using high-bandwidth NVLinks, as discussed in \sect{sect-2B}.

\begin{figure}[tb!] \centering
\begin{subfigure}[b]{0.8\columnwidth} \centering
		\includegraphics[width=\columnwidth]{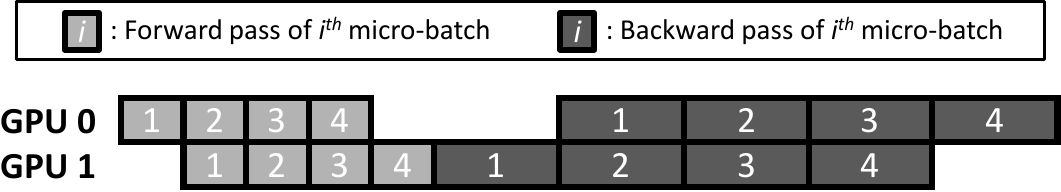}
		\vspace{-1.5em}
		\caption{GPipe scheduling}
		\vspace{0.5em}
\end{subfigure}
\begin{subfigure}[b]{0.8\columnwidth} \centering
		\includegraphics[width=\columnwidth]{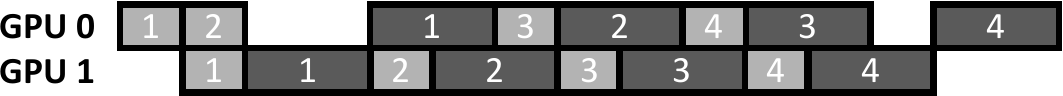}
		\vspace{-1.5em}
		\caption{1F1B scheduling}
\end{subfigure}
\caption{2-way pipeline parallelism using (a) GPipe and (b) 1F1B scheduling policy. Example assumes 4 micro-batches.}
\label{fig:pipeline_schedule}
\vspace{-1.2em}
\end{figure}

\textbf{Pipeline parallelism.} In pipeline parallelism, the communication always appears at the boundary of pipeline stages. \proposed models such behavior by inserting a P2P Send-Receive operator in between two consecutive stages (light gray operator in \fig{fig:tensor_and_pipeline}). Because \proposed supports both GPipe and 1F1B scheduling (\fig{fig:pipeline_schedule}), however, careful consideration must be made when determining the dependencies across different operators and its pipeline schedules. There are two types of dependencies to consider. First, the execution order within each GPU must be modeled. In GPipe, each GPU performs the forward pass for all micro-batches, followed by the backward pass. On the other hand, under 1F1B, each GPU must first fill in the pipeline and interleave the forward and backward passes for the remaining micro-batches. Along with the intra-GPU operator dependencies, the operators associated with the same micro-batch has to be strictly managed across the GPUs as well (e.g., forward pass of the $i^{th}$ micro-batch in GPU~2 cannot be executed earlier than that of GPU~1 in \fig{fig:pipeline_schedule}(b)). \proposed is designed to carefully enforce the unique scheduling constraints of GPipe and 1F1B, accurately modeling their inter-operator dependencies.

\subsection{Profiling Module}
\label{sect:profiler}
\begin{figure}[tb!] \centering
    \centerline{\includegraphics[width=\columnwidth]{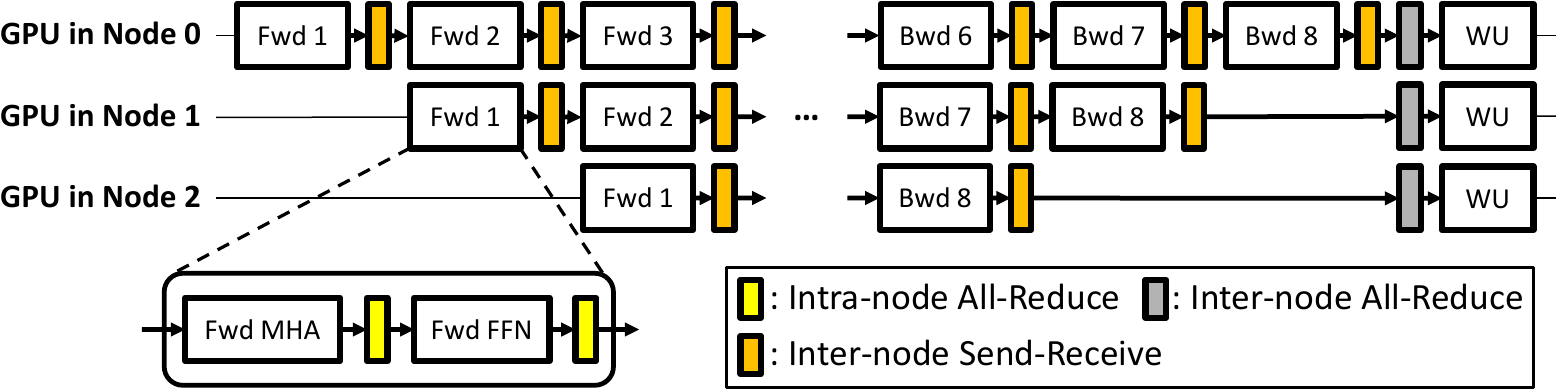}}
    \vspace{0.3em}
    \caption{An example illustrating how the 3D parallelism in  \fig{fig:3d_parallelism} (considering those 3 GPUs highlighted in red lines) is abstracted into an operator-granularity execution graph inserted with the required communication operators. "\texttt{Fwd i}" and "\texttt{Bwd i}" represent the $i^{th}$ micro-batch's forward and backward passes, respectively.}
    \label{fig:all_together}
    \vspace{-1.2em}
\end{figure}

\sethlcolor{SkyBlue}
The design objective of our profiling module is twofold: 1) determining the relationship between 
    the high-level layer-node's computation operator with the low-level CUDA kernels (tasks) and 
    2) profiling the execution time of each CUDA kernel over the target GPU device,
    which are utilized by \proposed's simulation algorithm (\sect{sect:sim_algo}) to estimate LLM's training time.
    The profiling module first executes all computation operators (e.g., forward pass of a MHA block)
    within the operator-granularity graph using the target GPU, one-by-one, and collects the CUDA kernel traces
    (e.g., \texttt{volta\_sgemm\_128x64\_tn} kernel for self-attention in MHA) at runtime using CUPTI
    (CUDA profiling tools interface)~\cite{cupti}.
    While CUPTI provides low-level information about the executed GPU kernels (e.g., names of the CUDA kernels and their execution time),
    it lacks the information regarding which operator each CUDA kernel is associated with.
    We employ the task-to-layer mapping methodology proposed by Zhu et al.~\cite{daydream} to identify 
    all the low-level CUDA kernels associated with a particular high-level graph operator.
    Using this information, vTrain constructs an \emph{operator-to-task lookup table} (\fig{fig:framework_overview}) 
    which provides both the list of CUDA kernels associated with a given operator and 
    the execution time of those kernels profiled over the target GPU.
A key challenge of profiling all the layer-node operators in the operator-granularity execution graph is its potentially high profiling overhead and slow simulation time. For an LLM with $L$ decoder layers trained with $N_{MB}$ micro-batches, the number of layer-nodes to profile and simulate becomes $O(L \times N_{MB})$. We make the key observation that practically all LLMs employ a model design methodology where an identically shaped decoder layer is stacked repeatedly with the same hyperparameters (\emph{h}, \emph{L}, \emph{s}, \emph{n}), each of which is partitioned evenly across all the GPUs for parallel training. This leads to each partitioned slice of the decoder layer, which is executed within a given GPU, to all consist of the same set of CUDA kernels, enabling us to drastically cut down on the set of operators to profile. Consider the operator-granularity execution graph in \fig{fig:all_together}. Because of LLM's repetitive execution nature of identical decoder layer operators, it is sufficient to profile only a single \texttt{Fwd MHA} layer-node instead of profiling all \texttt{Fwd MHA} layer-nodes in every \texttt{Fwd (i)}s. In the rest of this paper, we refer to these operations as \emph{necessary operators} (\fig{fig:framework_overview}). With such optimization in place, \proposed significantly reduces the number of operators to profile to $O(1)$. In \sect{sect:overhead}, we discuss \proposed's profiling overhead as well as its simulation time for deployment.

\subsection{(Low-level) Task-granularity Execution Graph}
\label{sect:task_exec_graph}
The operator-granularity execution graph effectively encapsulates the dependencies between computation and communication operators, but it can not infer how much latency is incurred to execute each operator. By referring to the operator-to-task lookup table, \proposed transforms the operator-granularity graph into a low-level task-level execution graph for accurate training time estimations. This is done by replacing each computation operators with a sequence of associated CUDA kernels (which is  listed precisely in the operator-to-task lookup table) while honoring inter-operator dependencies.

While these steps successfully transforms the computation operators into its corresponding CUDA kernels, \proposed still requires means to estimate the time taken to execute communication operators. 
\proposed employs a profiling-based approach for modeling intra-node communication where we measure the communication latency incurred across different communication data sizes and communication bandwidth over real multi-GPU systems. Specifically, we use NVIDIA's state-of-the-art NCCL library~\cite{nccl} to profile key communication primitives involved in 3D parallelism training (e.g., All-Reduce) over various data sizes and the number of GPUs involved in communication, which we utilize for estimating the latency of intra-node communication operators.
For modeling inter-node communications over multi-node distributed training systems, \proposed employs a latency-bandwidth model which utilizes the analytical model suggested by NVIDIA's NCCL to estimate All-Reduce communication latency of multi-node collective communication (detailed in \sect{sec-4}, \equat{eqn:comm}).

\begin{algorithm}[t!]
\footnotesize
\caption{Estimating an LLM's single-iteration training time using task-granularity execution graph}
\label{alg:simulation}
\renewcommand{\algorithmicrequire}{\textbf{Input:}}
\renewcommand{\algorithmicensure} {\textbf{Output:}}
\begin{algorithmic}[1]
\Require Task-granularity execution graph $G = \{ i: (V_i, E_i) \mid$ $V_i$ and $E_i$ are vertices and edges associated to GPU $i$, respectively $ \}$
\Ensure Iteration time $t_{iter}$
\State \textbf{Initialize} timeline $T$
\State \textbf{Initialize} task queue $Q \gets \emptyset$
\\
\ForAll{$i, (V_i, E_i) \in G$}
    \State $T[i] \gets 0$
    \State $Q \gets Q \cup \{ u | u \in V_i, u.ref = 0 \}$
\EndFor
\\
\While{$Q \neq \emptyset$}
    \State $u \gets Q.pop()$                                                \Comment{Fetch a task in FIFO order}
    \State $i \gets u.device$
    \State $T[i] \gets \max \left( T[i], u.start + u.duration \right)$      \Comment{Proceed the timeline}
    \ForAll{$c \in u.child$}
        \State $c.start \gets \max \left( c.start, u.start + u.duration \right)$
        \State $c.ref \gets c.ref - 1$
        \Comment{Update the child task}
        \If{$c.ref = 0$}
            \State $Q \gets Q \cup \{ c \}$                                  \Comment{Update the task queue}
        \EndIf
    \EndFor
\EndWhile
\\
\State $t_{iter} \gets \max \left( T[i] \right)$
\end{algorithmic}
\end{algorithm}

\subsection{Simulation Algorithm for Estimating Training Time }
\label{sect:sim_algo}

With the low-level task-granularity execution graph in place, \proposed now has means to estimate an LLM's single-iteration training time.
\algo{alg:simulation} provides a pseudo-code of how \proposed utilizes the task-level execution graph for estimating single-iteration training time.
The simulation begins by initializing the timeline and the task queue.
The timeline maintains the time advanced for each GPU within the training system.
Before the simulation begins, our simulator first searches all tasks in the task-granularity graph that have no dependencies,
    and places those tasks into the task queue (line $4$-$7$).
Afterwards, \proposed begins the simulation process to estimate training time, which goes through the following procedure:
   fetch a task from the task queue in FIFO order (line $10$), then advance the corresponding timeline accordingly (line $12$). 
 It is worth emphasizing that GPU kernel executions in LLM training are inherently sequential
    with few GPU kernels executed in parallel, due to the high computational intensity of these tasks.
Despite such property, our simulation must account for the possibility of computation-communication overlap,
    as illustrated in \fig{fig:framework_data_parallel}, which we faithfully model during the process in line $12$.
Following this, the states of the child tasks are updated (line $14$-$15$), and newly executable tasks are pushed into the task queue (line $16$-$18$).
The simulation is finished when there is no more tasks to execute at which point \proposed returns the predicted single-iteration training time.
The total end-to-end wall clock training time to consume all training tokens is estimated by multiplying ``predicted single-iteration training time''
    with ``the total number of training iterations an LLM training needs to go through (i.e., total number of tokens $/$ tokens consumed per iteration)''.

\subsection{Profiling Cost and Simulation Speed}
\label{sect:overhead}

Profiling a target LLM model configuration with hyperparameters (\emph{h}, \emph{L}, \emph{s}, \emph{n}) incurs $O(1)$ latency
    because we only need to profile the necessary operators (\sect{sect:profiler}).
This amounts to a simulation time of a single LLM training to be in the range of several seconds
    (e.g., $2$ seconds when evaluated over AMD EPYC 7502 32-Core CPU).
Consequently, conducting design space exploration to identify the optimal 3D parallelism strategy,
    which requires the evaluation of several thousands of different 3D parallelism's training time,
    takes only tens of minutes over a single CPU server as each simulation points are completely parallelizable over multiple CPU cores.
Overall, \proposed provides AI practitioners a highly productive toolset to develop efficient and cost-effective LLM training plans.

\section{Simulation and Validation Methodology} 
\label{sec-4}

We implemented \proposed on top of PyTorch 2.0.1~\cite{pytorch}.
The profiling module is implemented using CUPTI provided by NVIDIA CUDA Toolkit 11.8~\cite{cupti} which is integrated into PyTorch as a C++ extension for Python. 
To model LLM training, we leverage the code base of Microsoft's Megatron-DeepSpeed~\cite{megatron_deepspeed} which fully supports 3D parallelism for training transformer-based LLMs.
For validating \proposed's accuracy in predicting LLM's single-iteration training time, we separate our validation study into two parts, focusing on predicting the single-iteration training time over 1) a single-node containing $8$ GPUs and 2) multi-nodes containing 512 GPUs (all based on an NVIDIA A100 GPU).
Below we discuss the details of our \proposed validation methodology.

\begin{figure}[tb!] \centering
\hfill
\begin{subfigure}[b]{0.48\columnwidth}
    \includegraphics[width=\columnwidth]{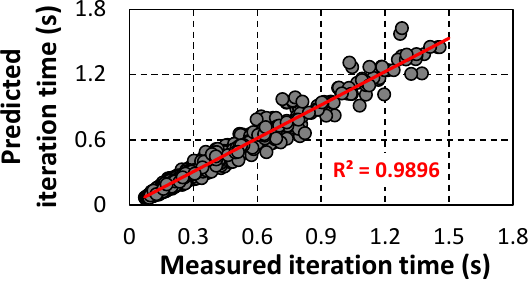}
    \vspace{-1.5em}
        \caption{Single-node validation}
\end{subfigure}
\begin{subfigure}[b]{0.48\columnwidth}
    \includegraphics[width=\columnwidth]{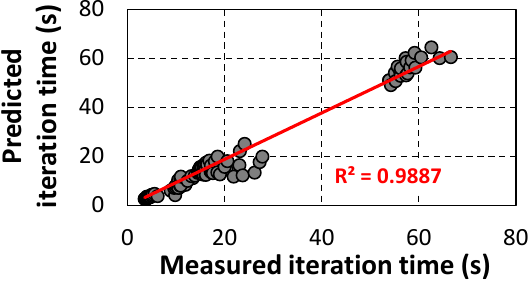}
    \vspace{-1.5em}
   \caption{Multi-node validation}
\end{subfigure}
\caption{Comparison of measured vs. estimated single-iteration LLM training time in (a) single-node training and (b) multi-node training. All training experiments are done using half (FP16) precision.}
\vspace{-1.5em}
\label{fig:validation_result}
\end{figure}

\textbf{Single-node training time validation.} For intra-node single-iteration training time validation, 
we perform the actual training of LLMs using a single AWS EC2 p4d GPU instance which contains 8 NVIDIA A100 GPUs connected over NVLink/NVSwitch~\cite{nvlink}.
We collected 1,440 data points with various LLM model configurations and tensor/data/pipeline parallelization plans. NVIDIA's NCCL~\cite{nccl} is used to profile and model key communication
primitives of LLM training whose communication data size ranges from 1 MB to 1024 MB.
The measured single-iteration training times of these data points over the 8 GPU system are
compared against the \proposed predicted training time, showing a mean absolute percentage error (MAPE) of 8.37\% with a coefficient of determination of 0.9896 (\fig{fig:validation_result}(a)).
One of the primary causes of vTrain's single-node simulation error is rooted in how
intra-node communication is modeled in our simulator. 
To model intra-node communication, 
        vTrain profiles All-Reduce latencies while running NCCL primitives 
        in an isolated environment.
     However, we observed   
that the latency of
        NCCL primitives during actual training is on average 30\% higher 
        than those measured in an isolated setting, a phenomenon also reported in \cite{daydream}.
Such discrepancy was especially more pronounced when tensor parallelism is employed.
    Specifically, \proposed underestimates the latencies of forward and backward passes of a tensor-parallel 
        transformer layer 
        because tensor parallelism exhibits more frequent All-Reduce operations
        (twice for each forward and backward passes of a transformer layer) than
        other parallelism strategies. 
        Further improving \proposed to better accommodate such behavior for intra-node communication is left as future work.

\textbf{Multi-node training time validation.} Similar to single-node validation, we perform the actual training of various LLM model configurations and measure their single-iteration training times for our multi-node training validation. 
Our multi-node GPU system contains 64 multi-GPU server nodes where each node contains $8$ NVIDIA A100 GPUs, amounting to 512 GPUs. Intra-node communication is done over NVLink/NVSwitch while inter-node communication is conducted over four 200 Gbps Mellanox HDR InfiniBand HCAs. The server nodes are connected in a two-level non-blocking fat-tree topology. Due to the enormous monetary cost associated with LLM training at this scale, collecting sufficient amount of validation data for academic research in itself is an extremely challenging task -- to the best of our knowledge, the only publicly available real-world measurements of large-scale LLM training (wall clock) time is reported in Microsoft-NVIDIA's work on Megatron-Turing NLG~\cite{mt_nlg} which only includes ``3'' LLM training data points.
With the help of our collaborators in industry, we were able to secure $116$ data points for validating the aforementioned multi-node training time (i.e., total of $512$ A100 GPUs), which we compared against the \proposed-predicted training time.
The validation data points were collected based on the LLM model configurations of Megatron-LM~\cite{megatron2}.
As discussed in \sect{sect:task_exec_graph}, we use a simple latency-bandwidth model to estimate inter-node communication time.
Specifically, we calculate the communication latency of each All-Reduce operation (the most critical collective communication primitive in 3D parallel training, \fig{fig:framework_data_parallel}) using the following analytical model based formula suggested in NVIDIA's NCCL~\cite{nccl-test}:

%\footnote{\todo{To foster future works in this research space, we released all of our validation data to the public.}}

\begin{equation}
\label{eqn:comm}
    t = \frac{S}{B} \cdot \frac{2(n-1)}{n}
\end{equation}

Here $t$ is the derived communication time, $S$ is the size of data being transferred, $n$ is the number of workers (i.e., GPUs) participating in the collective communication, and $B$ is the \emph{effective} communication bandwidth. 
To better model the effective communication bandwidth, we introduce a tuning parameter called \emph{bandwidth effectiveness factor}, denoted as $\alpha$, and express the effective communication bandwidth as $B=\alpha B_{max}$ where $B_{max}$ is the maximum interconnection bandwidth, i.e., 800 Gbps ($=$ 200 Gbps $\times$ 4). 
To identify the optimal $\alpha$ value, we sweep this value from 0.1 to 1.0 and compare the \proposed-predicted single-iteration training times vs. the measured single-iteration training time. We observed that the average prediction error was minimized at $\alpha=1.0$ in our evaluation setting, which assumes that all the inter-node network bandwidth can fully be utilized. Using an $\alpha$ value of 1.0, our multi-node training validation study showed a MAPE of 14.73\% with a coefficient of determination of 0.9887  (\fig{fig:validation_result}(b)). 
The primary causes of vTrain's multi-node simulation error are rooted on our simple latency-bandwidth model utilized for estimating inter-node communication latency, which introduces two main avenues of improvement. First, our inter-node communication model does not capture the effect of straggler GPU node’s training time at synchronization points, nor does it account for the latency overheads of NCCL kernel launches. Second, it is challenging for the latency-bandwidth communication model to accommodate the dynamic behaviors of a large, complicated network topology. For example, in \fig{fig:3d_parallelism}, there are four different data
    parallel groups across Node 0 and Node 3.
Each group performs All-Reduce communications, and these communications from the 
    four groups (depicted as gray arrows) share the same network building blocks
    (e.g., ToR switches), which can lead to potential interference between 
    the groups.
However, our analytical model is not able to capture such dynamic effects 
    as it only accepts statically determined data size, number of GPUs involved in the inter-node communication, 
    and link bandwidth as inputs.
We believe the simulation errors of vTrain's multi-node training time can be alleviated 
    by incorporating the dynamic nature of inter-node communication 
    into our analytical model or by employing existing sophisticated
    modeling techniques~\cite{astra-sim-2, calculon}. We leave the augmentation of such advanced communication model as future work.

\begin{table*}[thb!]
    \hfill
    \caption{Comparison of baseline MT-NLG training plans vs. \proposed uncovered, more cost-effective training plans (``Our findings'').}
    \vspace{0.6em}
    \label{tab:mtnlg}
    \resizebox{\textwidth}{!}{%
    \begin{tabular}{|ccccccc||ccccccc|}
    \hline
    \multicolumn{7}{|c||}{\textbf{MT-NLG}} &
    \multicolumn{7}{c|}{\textbf{Our findings}} \\ \hline
    \multicolumn{1}{|l|}{$(t, d, p)$} &
    \multicolumn{1}{l|}{\begin{tabular}[c]{@{}l@{}}Iteration\\ time (s)\end{tabular}} &
    \multicolumn{1}{l|}{\begin{tabular}[c]{@{}l@{}}\textbf{Total training} \\ \textbf{time (days)}\end{tabular}} &
    \multicolumn{1}{l|}{\begin{tabular}[c]{@{}l@{}}GPU compute\\ utilization (\%)\end{tabular}} &
    \multicolumn{1}{l|}{\# GPUs} &
    \multicolumn{1}{l|}{\$ per hour} &
    \begin{tabular}[c]{@{}l@{}}\textbf{\$ in total}\\ \textbf{(million)}\end{tabular} &
    \multicolumn{1}{l|}{$(t, d, p)$} &
    \multicolumn{1}{l|}{\begin{tabular}[c]{@{}l@{}}Iteration\\ time (s)\end{tabular}} &
    \multicolumn{1}{l|}{\begin{tabular}[c]{@{}l@{}}\textbf{Total training}\\ \textbf{time (days)}\end{tabular}} &
    \multicolumn{1}{l|}{\begin{tabular}[c]{@{}l@{}}GPU compute\\ utilization (\%)\end{tabular}} &
    \multicolumn{1}{l|}{\# GPUs} &
    \multicolumn{1}{l|}{\$ per hour} &
    \begin{tabular}[c]{@{}l@{}}\textbf{\$ in total}\\ \textbf{(million)}\end{tabular} \\ \hline
    \multicolumn{1}{|c|}{(8, 8, 35)} &
    \multicolumn{1}{c|}{42.59} &
    \multicolumn{1}{c|}{\textbf{33.52}} &
    \multicolumn{1}{c|}{42.67} &
    \multicolumn{1}{c|}{2,240} &
    \multicolumn{1}{c|}{11,200} &
    \textbf{9.01} &
    \multicolumn{1}{c|}{(8, 12, 21)} &
    \multicolumn{1}{c|}{45.29} &
    \multicolumn{1}{c|}{\textbf{35.64}} &
    \multicolumn{1}{c|}{44.58} &
    \multicolumn{1}{c|}{2,016} &
    \multicolumn{1}{c|}{10,080} &
    \textbf{8.62} \\ \hline
    \multicolumn{1}{|c|}{(8, 10, 35)} &
    \multicolumn{1}{c|}{34.92} &
    \multicolumn{1}{c|}{\textbf{27.49}} &
    \multicolumn{1}{c|}{41.63} &
    \multicolumn{1}{c|}{2,800} &
    \multicolumn{1}{c|}{14,000} &
    \textbf{9.24} &
    \multicolumn{1}{c|}{(8, 16, 21)} &
    \multicolumn{1}{c|}{34.97} &
    \multicolumn{1}{c|}{\textbf{27.53}} &
    \multicolumn{1}{c|}{43.30} &
    \multicolumn{1}{c|}{2,688} &
    \multicolumn{1}{c|}{13,440} &
    \textbf{8.88} \\ \hline
    \multicolumn{1}{|c|}{(8, 12, 35)} &
    \multicolumn{1}{c|}{29.81} &
    \multicolumn{1}{c|}{\textbf{23.46}} &
    \multicolumn{1}{c|}{40.64} &
    \multicolumn{1}{c|}{3,360} &
    \multicolumn{1}{c|}{16,800} &
    \textbf{9.46} &
    \multicolumn{1}{c|}{(8, 20, 21)} &
    \multicolumn{1}{c|}{28.78} &
    \multicolumn{1}{c|}{\textbf{22.65}} &
    \multicolumn{1}{c|}{42.09} &
    \multicolumn{1}{c|}{3,360} &
    \multicolumn{1}{c|}{16,800} &
    \textbf{9.13} \\ \hline
    \end{tabular}%
    }
\end{table*}

\begin{figure}[tb!] \centering
    \begin{subfigure}[b]{0.48\columnwidth}
        \includegraphics[width=\columnwidth]{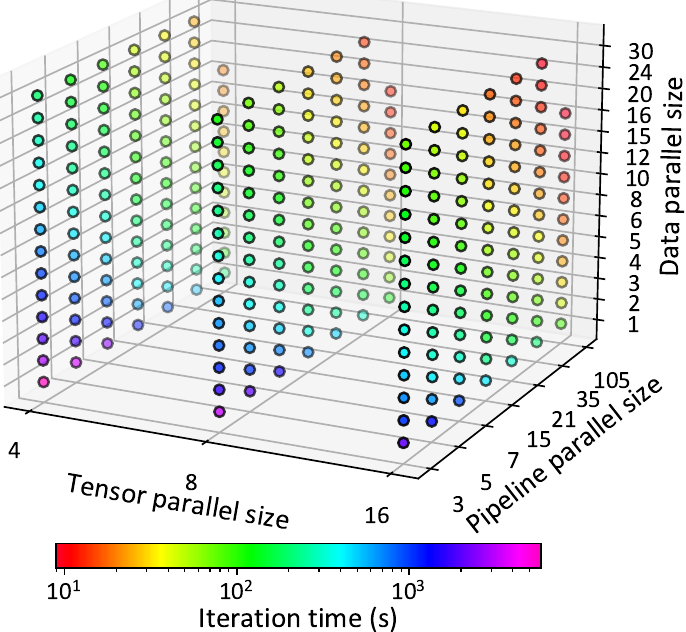}
        \vspace{-1.3em}
        \caption{}
    \end{subfigure}
    \hspace{0.1em}
    \begin{subfigure}[b]{0.48\columnwidth}
        \includegraphics[width=\columnwidth]{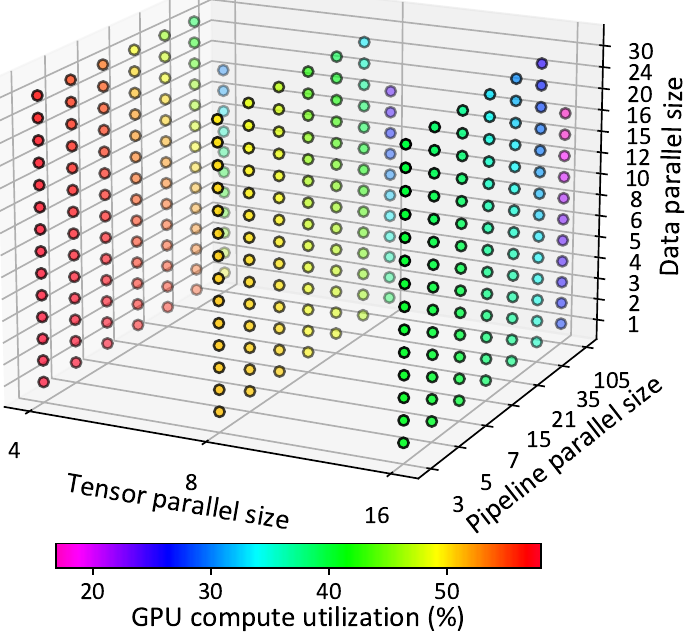}
        \vspace{-1.3em}
        \caption{}
    \label{fig:mtnlg_util}
    \end{subfigure}
    \caption{Design space exploration using \proposed to predict MT-NLG's (a) single-iteration training time and (b) GPU compute utilization.}
    \vspace{-0.5em}
    \label{fig:mtnlg_exploration}
\end{figure}

\section{Design Space Exploration for Cost-Effective and Compute-Optimal LLM Training using \proposed}
\label{sect:case_studies}

We now turn our attention to studying how \proposed can be utilized to determine an efficient LLM parallelization strategy
    and the training system configuration for cost-efficient and compute-optimal LLM training. 

\subsection{Case Study \#1: Cost-effective LLM Training Plan}
\label{sect:cost-optimal-case-study}

As discussed in \fig{fig:compute_util_to_training_time}, a decrease in GPU compute utilization directly translates into a significant increase in training cost. 
Given a target LLM, the available GPU compute budget, and the training token size, our first case study seeks to answer the following research question:
    what is the most \emph{cost-effective} training parallelization strategy and AI training system configuration
    that effectively balances end-to-end training time \emph{and} its associated training cost?

We use the state-of-the-art $530$B Megatron-Turing NLG (MT-NLG)~\cite{mt_nlg} model architecture and its parallelization strategy 
    as a representative example to demonstrate \proposed's feasibility in rapidly identifying an LLM training plan with optimal cost-efficiency.
We focus on MT-NLG because it is most open to providing details on the hyperparameters of its training system configuration.
Nonetheless, \proposed's simulation methodology is equally applicable to other LLM training systems employing 3D parallelism.
Under the baseline MT-NLG study, the authors employ 3D parallelism targeting the following LLM training scenario:

\begin{itemize}
    \item ({\bf Model}) A decoder-only LLM with 20,480 of hidden size (\emph{h}), 105 decoder layers (\emph{L}), and 128 attention heads (\emph{n}).
    \item ({\bf Compute budget}) MT-NLG presents three data points employing (8, \emph{d}, 35)-way 3D parallelismwith \emph{d}$=$$8$-/$10$-/$12$-way data parallelism,
            amounting to a compute budget of ($8 \times d \times 35$) $=$ ($280 \times d$, $d$ = 8, 10, 12) GPUs.
    \item ({\bf Training dataset}) A total of $270$ billion training tokens where each training iteration consumes a  batch size of (1,920 $\times$ 2,048) tokens
            across all the GPUs in the system, leading to a total of approximately 68,000 training iterations for end-to-end training. 
\end{itemize}

In order to train MT-NLG~\cite{mt_nlg}, the authors present three heuristic based LLM training plans they employed 
    to balance training throughput and overall cost (the left half of \tab{tab:mtnlg}).
Identifying a cost-effective LLM training plan requires countless trial-and-error exploration of various training plans in practice 
    which, due to the enormous cost associated with such design space exploration, can only cover a subset of the full design space of 3D parallelism. 
In this section, we utilize \proposed to perform a \emph{full} design space exploration of (\emph{t}, \emph{d}, \emph{p})-way 3D parallelism
    by sweeping through the design space up to \emph{t$_{max}$}=16, \emph{d$_{max}$}=32, and \emph{p$_{max}$}=105-way tensor/data/pipeline parallelism. 
\fig{fig:mtnlg_exploration} illustrates the result of such design space exploration in terms of MT-NLG model's
    (a) single-iteration training time and (b) its GPU compute utilization. 
Not surprisingly, performance is best when the LLM training system employs a large number of GPUs, 
    e.g., (16, 16, 105)-way 3D parallelism (\fig{fig:mtnlg_exploration}(a)). 
Unfortunately, such design point requires roughly $10\times$ more GPUs than the baseline MT-NLG
    and also suffers from significant GPU compute underutilization (average $17\%$ GPU utilization, \fig{fig:mtnlg_exploration}(b)),
    worsening training cost-efficiency than all the baseline MT-NLG design points.
Consequently, finding the most cost-optimal LLM training plan requires careful consideration of both training iteration time \emph{and} GPU compute utilization.

\begin{figure}[tb!] \centering
    \includegraphics[width=0.9\columnwidth]{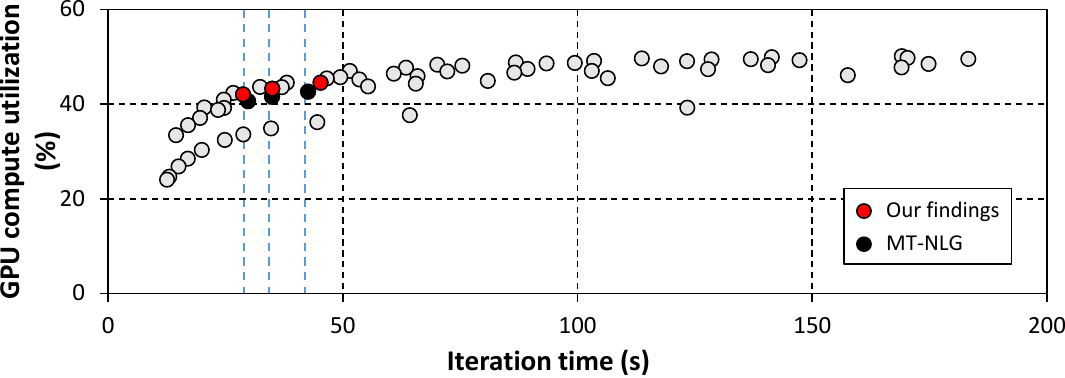}
    \vspace{0.5em}
    \caption{
        In this figure, we select design points in \fig{fig:mtnlg_exploration} that employ $8$-way tensor parallelism
            to re-plot the relationship between training iteration time and GPU compute utilization.
        The black dots represent the three design points proposed in the original MT-NLG study 
            and the three red dots represent more cost-effective design points uncovered with \proposed.
    }
    \vspace{-0.8em}
    \label{fig:mtnlg_exploration_2}
\end{figure}

In \fig{fig:mtnlg_exploration_2}, we re-plot the design space exploration results in \fig{fig:mtnlg_exploration} 
    by limiting the design points to those employing $8$-way tensor parallelism 
    and summarize its single-iteration training time and the resulting GPU compute utilization.
Since the baseline MT-NLG explores three design points with a total of 2,240/2,800/3,360 GPUs, 
    we focus on those design points that require similar amount of GPU compute budget for a fair comparison. 
The results in \fig{fig:mtnlg_exploration_2} show that \proposed can effectively uncover LLM training plans (the three red dots) 
    that are competitive to the baseline MT-NLG training systems (the three black dots).
We summarize such results in \tab{tab:mtnlg} where each row represents a given design point (1) proposed in MT-NLG and 
    (2) the \proposed uncovered, cost-effective LLM training system configurations (which was identified within several minutes).
For instance, the first row in \tab{tab:mtnlg} compares the baseline MT-NLG vs. \proposed recommended designs, each requiring 2,240 GPUs vs. 2,016 GPUs, respectively.
Notice how the design points uncovered with \proposed requires $10\%$ fewer GPUs, achieves $4.5\%$ higher GPU utilization,
    but only incurs $6.3\%$ longer end-to-end training time (i.e., additional two days).
Using AWS EC2 P4d GPU instances~\cite{amazon_p4d} as a proxy to translate these numbers in monetary costs,
    our \proposed recommended design effectively reduces the training cost by $5\%$, from 9.01 to 8.62 million dollars, saving 0.39 million dollars. 
More critically, \proposed was able to identify such competitive LLM training plan 
    within several minutes (i.e., exploring the full design space of \fig{fig:mtnlg_exploration} takes less than 200 seconds),
    demonstrating the benefits our fast simulation framework provides to practitioners seeking to determine an efficient LLM parallelization
    and training plan across a distributed GPU system.

\begin{table}[t!]
    \centering
    \vspace{0.7em}
    \caption{
        Validation of \proposed-predicted single-iteration training time vs. its actual single-iteration training time measured 
            over our real $64$/$256$/$512$ A100 multi-GPU systems. $m$ refers to the micro-batch size.
        The design points highlighted in \textbf{bold} are LLM training plans newly uncovered by \proposed's full design space exploration 
            which achieve optimal cost-effectiveness (i.e., equivalent to the ``Our findings'' column in \tab{tab:mtnlg}), 
            consistently providing lower single-iteration training time across all GPU sizes 
            (the percentage of training time reduction is denoted inside the parentheses).
    }
    \label{tab:mtnlg_small_scale}
    \resizebox{\columnwidth}{!}{%
    \begin{tabular}{|c|c|c|c|c|c|}
        \hline
        \begin{tabular}[c]{@{}c@{}}Param.\\ (B)\end{tabular} &
        \# GPUs & &
        $(t, d, p, m)$ &
        \begin{tabular}[c]{@{}c@{}}Predicted\\ iteration time (sec)\end{tabular} &
        \begin{tabular}[c]{@{}c@{}}Measured\\ iteration time (sec)\end{tabular} \\ \hline
        \multirow{2}{*}{3.6}  & \multirow{2}{*}{64}  & \cite{megatron2} & (2, 32, 1, 16) & 2.919                   & 3.938                   \\ \cline{3-6} 
                              &                      & Ours & \bf{(1, 64, 1, 8)}  & \bf{2.746 (6\%$\downarrow$)}  & \bf{3.567 (9\%$\downarrow$)}  \\ \hline
        \multirow{2}{*}{18.4} & \multirow{2}{*}{256} & \cite{megatron2} & (8, 32, 1, 4)  & 7.533                   & 9.928                   \\ \cline{3-6} 
                              &                      & Ours & \bf{(8, 32, 1, 8)}  & \bf{7.259 (4\%$\downarrow$)}  & \bf{9.604 (3\%$\downarrow$)}  \\ \hline
        \multirow{2}{*}{39.1} & \multirow{2}{*}{512} & \cite{megatron2} & (8, 32, 2, 4)  & 13.859                  & 14.757                  \\ \cline{3-6} 
                              &                      & Ours & \bf{(4, 32, 4, 2)}  & \bf{12.226 (12\%$\downarrow$)} & \bf{13.876 (6\%$\downarrow$)} \\ \hline
        \end{tabular}%
    }
    \vspace{-0.8em}
\end{table}

Because the LLM training costs reported in \tab{tab:mtnlg} are based on \proposed's ``simulated'' single-iteration training time,
    we now quantitatively demonstrate how effective \proposed is in accurately estimating efficient training plans,
    cross-validating the vTrain simulated training time vs. the actual training time measured over real multi-node GPU systems.
Validating the results of our design space exploration in \fig{fig:mtnlg_exploration} and \tab{tab:mtnlg} over several thousands of GPUs is 
    practically challenging for academic researchers due to the astronomical costs associated with such study.
We therefore perform a design space exploration over a scaled-down version of MT-NLG that can be trained under 512 GPUs, the methodology of which we detailed below.
In~\cite{megatron2}, the authors evaluate several scale-down versions of MT-NLG using 3D parallelism over distributed A100 GPU systems.
These scale-down versions of MT-NLG maintain the identical, decoder-only transformer architecture but with reduced hyperparameters, 
    such as smaller hidden size and reduced number of layers.
Using these design points presented in~\cite{megatron2}, we take the following measure to validate 
    how effective \proposed is in estimating efficient LLM training plans, the result of which is summarized in \tab{tab:mtnlg_small_scale}.
\tab{tab:mtnlg_small_scale} shows \emph{two} LLM training plans (1) those that are employed in~\cite{megatron2} to train MT-NLG over $64$/$256$/$512$ A100 GPU systems,
    and (2) those that \proposed's full design space exploration uncovered as providing better cost-effectiveness.
The single iteration training time of these two LLM training plans (``\cite{megatron2}'' and ``Ours'' in \tab{tab:mtnlg_small_scale})
    are then evaluated using both \proposed (``Predicted'') and real (``Measured'') multi-node GPU systems.
As depicted in \tab{tab:mtnlg_small_scale}, across all studied (64/256/512) GPU system configurations,
    the LLM training plan suggested by \proposed is able to consistently outperform those reported in~\cite{megatron2},
    for ``both'' predicted and real training time measurements.
These results highlight \proposed's fidelity as well as its usefulness in cost-efficiently exploring the design space of LLM training plans in a fast and accurate manner.

\subsection{Case Study \#2: Cost-effective Multi-tenant GPU Clusters}
\label{sect:cluster}

The previous case study in \sect{sect:cost-optimal-case-study} assumed a scenario 
    where a \emph{single} large-scale training job occupiesthe majority of GPU cluster's usage to train an LLM
    (e.g., jobs that train a \emph{foundation model} from scratch~\cite{gpt3,gpt4,opt,llama,llama2}),
    one which \proposed can provide valuable insights in determining what is the most cost-effective LLM training plan.
With the proliferation of open-source LLMs like OPT~\cite{opt} or LLaMA~\cite{llama, llama2},
    ML researchers are putting significant efforts in fine-tuning these foundation models to optimize them for a specific application domain
    (e.g., chatbots~\cite{chatgpt} and text-to-code~\cite{openai_codex, github_copilot}).
As such, major cloud service providers are now offering services that ease the process of fine-tuning generative AI models
    (e.g., Google Cloud's Generative AI Studio~\cite{genai_studio} or Microsoft Azure's OpenAI Service~\cite{openai_service}).
Given this landscape, another common LLM training scenario is where \emph{multiple} LLM training jobs 
    are simultaneously being scheduled and executed over a ``multi-tenant'' GPU cluster containing tens of thousands of GPUs.
From the cloud service provider's perspective, maintaining high service quality to end-users (e.g., minimizing job completion time and makespan)
    while also maximizing the utilization of GPU clusters is critical to optimize total cost of ownership (TCO). 

There exists a long list of prior work seeking to optimize the aforementioned design objectives (i.e., maximize GPU utilization and service quality) for large-scale
    multi-tenant GPU cluster scheduling~\cite{gandiva,heteroaware_cluster,chronus,tiresias,themis_nsdi,optimus,pollux,gandiva_fair,zhao22,elasticflow,lucid,bian21,hu21,antman,jeon19,hwang2021,sia}.
State-of-the-art solutions in this research space are designed towards satisfying the following properties:

\begin{itemize}
\item ({\bf Serverless training}) ML researchers typically have limited expertise in low-level \emph{systems} issues.
        As such, it is extremely valuable to free these researchers from painstakingly configuring the training system parameters
            (e.g., number of GPUs to utilize for training, model parallelization plans, making sure the overall memory usage fits within the GPU memory, etc)
            and allowing them to focus on the algorithmic aspects of ML.
        Recent work~\cite{elasticflow,pollux,optimus,themis_nsdi,lucid,sia,antman} thus employ a \emph{serverless} computing model
            where each ML training job leaves the systems issue up to the GPU cluster manager and only specifies the ML model it wants to train
            (e.g., model architecture, training hyperparameters like learning rate)
            and what is the expected training performance guarantees it expects
            (e.g., guaranteeing to finish a job before a given deadline).
\item ({\bf Elastic resource scaling}) With such serverless computing model in place,
            an important research problem to address from the perspective of GPU cluster management is
            how to intelligently allocate GPU resources to individual training jobs that satisfy various design objectives.
        As such, \emph{elastic} GPU resource scaling policies have been developed that try to balance GPU utilization
            and service quality~\cite{elasticflow,pollux,themis_nsdi,optimus,hwang2021,sia,gandiva}.
        Among these, ElasticFlow~\cite{elasticflow} is a state-of-the-art solution that proposes to \emph{profile} the training job's performance offline
            and utilize that information for elastic GPU resource scaling and deadline-aware job scheduling.
        Specifically, ElasticFlow profiles a target ML model's parallelization plan by measuring how the training performance scales
            as a function of the number of GPUs utilized for training.
        ElasticFlow, however, limits its parallelization space exploration only to data parallelism (see \sect{sect-2B})
            leading to sub-optimal scheduling decisions.
        In this case study, we utilize \proposed's ability to identify optimal LLM training plans (as discussed in \sect{sect:cost-optimal-case-study})
            and how it can help optimize multi-tenant GPU scheduling choices.
\end{itemize}

We employ a simulation-based methodology to demonstrate \proposed's effectiveness
    because the estimated cost of running the same experiments we present in this section at Amazon AWS
    is over 100 million dollars (we study a total of $41$ workload traces,
    each trace modeling a GPU cluster operating for 400 hours of wall clock training time using 1,024 A100 GPUs)
    \footnote{It is worth emphasizing that ElasticFlow, due to similar cost reasons, demonstrated their proposal's efficacy using simulation.}.
We assume a GPU cluster with 128 nodes (each node containing $8$ A100 GPUs, identical to our multi-node validation study in \sect{sec-4}), amounting to 1,024 GPUs in total. 
Our simulator evaluates the entire lifetime of a training job, from its arrival to its completion.
In terms of the workloads we evaluate, we use real-world traces from Microsoft's internal ITP clusters~\cite{ms_cluster_trace}
    to model the arrival time of each training job.
Specifically, a single workload trace is generated by randomly sampling $N$ consecutive job arrival points
    from Microsoft's trace which are used to model the time intervals in which two consecutive jobs arrive to the training cluster.
All workload traces are assumed to arrive within a fixed time period,
    meaning a workload traces with larger number of jobs will likely incur larger deadline violation rates
    (i.e., workload trace with $128$ jobs will stress the training cluster more than $64$ jobs thereby invoking larger number of deadline violations).
Each arrived job is randomly chosen among the three LLM configurations listed in \tab{tab:cluster_models}
    whose total number of training iterations to execute and its requested deadline guarantee is also randomly chosen.

\begin{table}[]
    \vspace{0.7em}
    \caption{
        Model configurations and batch sizes used in our GPU cluster experiments in \sect{sect:cluster}.
        $L$ is the number of transformer layers, $h$ is the size of hidden dimensions, $n$ is the number of attention heads,
            $s$ is the maximum sequence length, and $B$ is the batch size used for the LLM training job.
        }
    \label{tab:cluster_models}
    \centering{\footnotesize %
    \begin{tabular}{|c|c|c|c|c|c|}
    \hline
    \textbf{Parameters (billion)} & \boldmath$L$ & \boldmath$h$ & \boldmath$n$ & \boldmath$s$ & \boldmath$B$ \\ \hline
        18.4 & 40 &  6,144 & 48 & 2,048 & 1,024 \\ \hline
        39.1 & 48 &  8,192 & 64 & 2,048 & 1,536 \\ \hline
        81.2 & 64 & 10,240 & 80 & 2,048 & 1,792 \\ \hline
    \end{tabular}%
    }
\end{table}

We implement the exact same scheduling algorithm ElasticFlow proposes,
    which elastically adjusts the GPU resources allocated to each training job based on the profiled training performance numbers.
The difference in how well baseline ElasticFlow vs. \proposed optimize scheduling decisions primarily lies in
    how close the best profiled training performance is to the performance achievable with an optimal parallelization plan.
For baseline ElasticFlow, limiting its profiled parallelization plans only to the data parallelism dimension
    while fixing the tensor/pipeline parallelism dimension value to ``one'' (which is what ElasticFlow assumes in \cite{elasticflow})
    severely constrains the possible design points the GPU cluster scheduler can leverage for scheduling decisions.
We therefore have baseline ElasticFlow keep a \emph{minimum} tensor/pipeline parallelism degree necessary for each LLM model configuration,
    e.g., the baseline ElasticFlow policy parallelizes the $39.1$B LLM model in \tab{tab:cluster_models} across ($16$$\times$$d$) GPUs
    using 8-way tensor parallelism, 2-way pipeline parallelism, and $d$-way data parallelism.
Our \proposed, on the other hand, can utilize its knowledge of the optimal parallelization plan for scheduling,
    guaranteeing at a minimum to provide the \emph{same} training performance that baseline ElasticFlow can provide.
Below we detail the effectiveness of \proposed in terms of deadline satisfactory rate, job completion time, and makespan.

\begin{figure}[tb!] \raggedleft
\begin{subfigure}[b]{\columnwidth}
    \includegraphics[width=0.94\columnwidth]{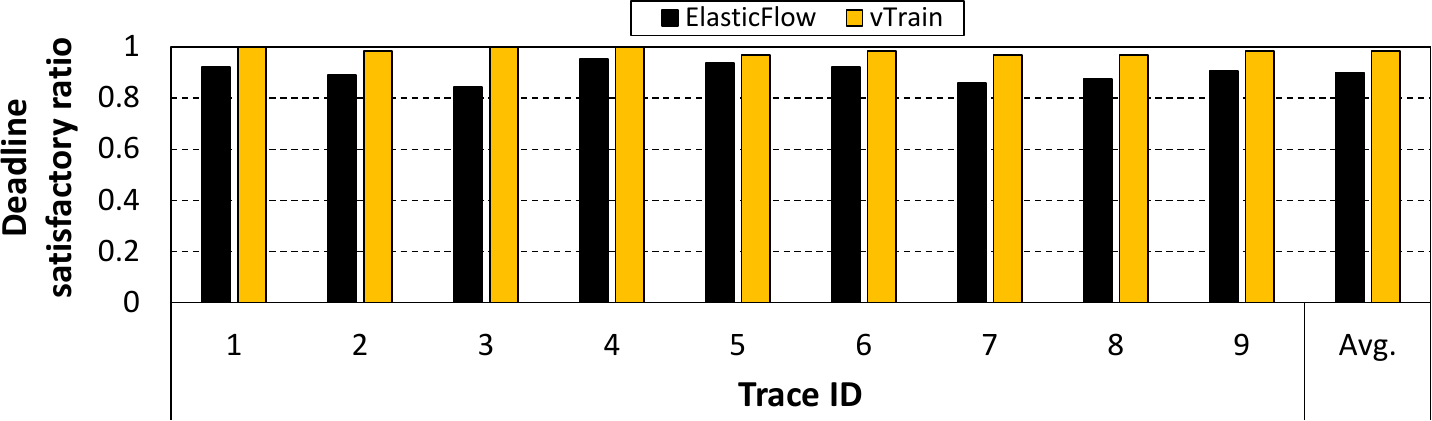}
    \vspace{-0.3em}   
    \caption{64 jobs}
    \vspace{0.5em}
\end{subfigure}
\begin{subfigure}[b]{\columnwidth}
    \includegraphics[width=0.94\columnwidth]{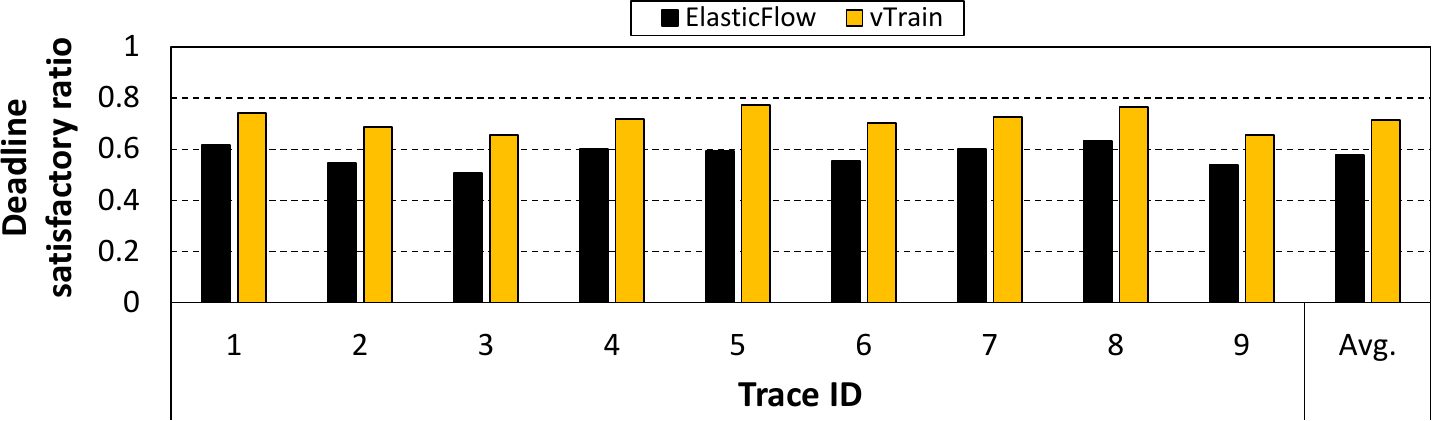}
    \vspace{-0.3em}
    \caption{128 jobs}
\end{subfigure}
\caption{Comparison of deadline satisfactory ratio over ten workload traces.
        Traces with $128$ jobs stresses the GPU cluster more than $64$ jobs and lead to higher deadline violations.}
\label{fig:deadline}
\end{figure}

\textbf{Deadline satisfactory ratio} is defined as the ratio of jobs that have met their deadlines,
    one which is used as the most important metric to highlight ElasticFlow's deadline-aware scheduling algorithm. 
To evaluate this metric, we set the deadline for each job as $\lambda \cdot \textit{duration}$ after its arrival,
    where $\lambda$ is sampled from a uniform distribution [0.5, 1.5].
We simulate 64-job and 128-job workloads synthesized from each trace for both baseline and \proposed-enabled systems and measure the deadline satisfactory ratio.
As shown in \fig{fig:deadline}, the \proposed-enabled system always satisfies more job requests than the baseline ElasticFlow,
    improving the deadline satisfactory ratio compared to the baseline system by 1.09$\times$ and 1.23$\times$ on average for 64-job and 128-job workloads, respectively.

\begin{figure}[t] \centering
    \includegraphics[width=0.94\columnwidth]{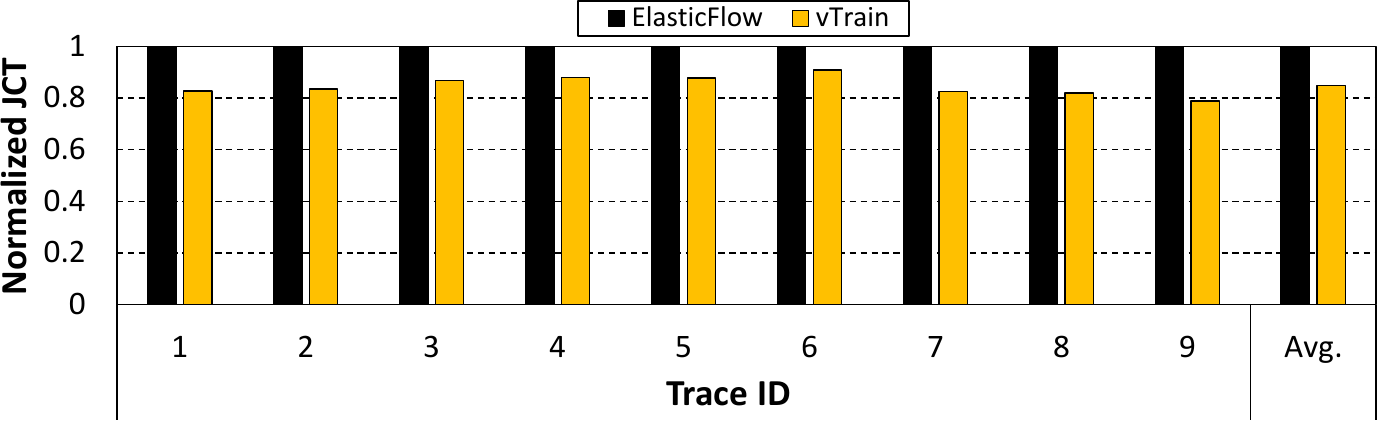}
    \vspace{-0.3em}
    \caption{Comparison of average job completion time (32 jobs). Results are normalized to ElasticFlow.}
    \label{fig:jct}
\end{figure}

\textbf{Job completion time (JCT)} of a job is the time spent from its arrival to its completion. 
Since ElasticFlow simply terminates jobs that cannot be finished within their deadlines (which can artificially lower JCT),
    deriving JCT under a deadline-free setting is another important metric to measure for a fair comparison.
Thus, we generate nine 32-job traces (without specifying job deadlines) which has a relatively lenient traffic condition. As shown in \fig{fig:jct}, 
    \proposed reduces JCT by an average 15.21\% while always guaranteeing JCT to become shorter than ElasticFlow.

\begin{figure}[tb!] \centering
    \includegraphics[width=0.8\columnwidth]{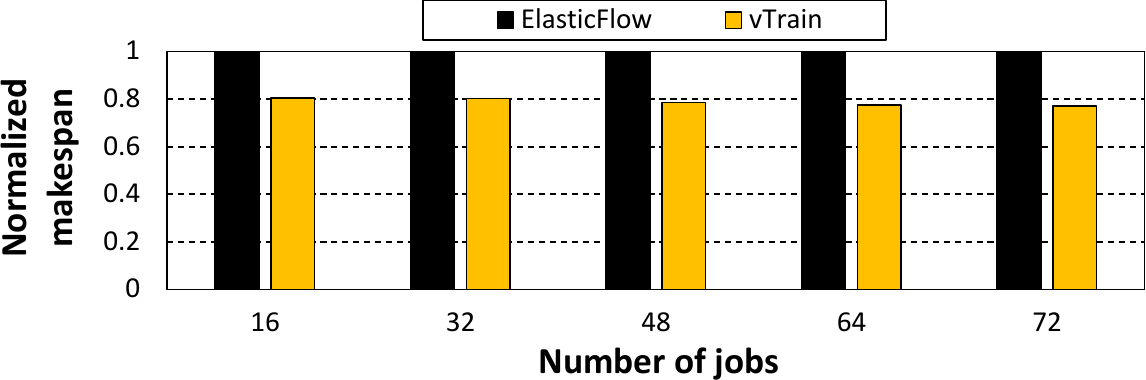}
    \caption{Comparison of makespan (normalized to ElasticFlow).}
    \label{fig:makespan}
    \vspace{-0.8em}
\end{figure}

\textbf{Makespan} is the time spent until all jobs are completely executed.
For the purpose of this study, we generate five synthetic workload traces (containing $16$ to $72$ jobs)
    where the $N$ jobs are all submitted to the GPU cluster at the same time.
We then measure the makespan for both systems, the result of which is summarized in \fig{fig:makespan}.
For the 16-job workload which represents the lightest traffic, ElasticFlow shows the least difference in makespan vs. \proposed. 
As the number of jobs increases, however, baseline suffers from effectively allocating limited GPU resources to the jobs
    due to its sub-optimal choice of parallelization plans.
Our \proposed helps the cluster scheduling algorithm better utilize the scarce GPU resources and reduces makespan up to 23.03\%.

\subsection{Case Study \#3: Compute-optimal LLM Model Size}
\label{sect:chinchilla}

The Chinchilla scaling law~\cite{chinchilla} empowers AI practitioners to discover the largest model and maximum number of tokens
    that can be trained within a given compute budget (\sect{sect2A}).
Since the accuracy of an AI model heavily depends on the model parameter size and the amount of data used for training,
    understanding the relationship between compute budget, largest trainable model size,
    and maximum token counts to train (which we refer to as compute-optimal model/token size) becomes critical for AI practitioners.

Unfortunately, na\"ively assuming that the available compute budget is singlehandedly determined by how many GPUs one can utilize will lead to misleading conclusions.
As shown in \fig{fig:mtnlg_exploration}(b), GPU utilization varies significantly depending on the parallelization strategy employed by the training system,
    leading to large variation in \emph{effective} FLOPS the LLM can be trained on.
Therefore, it is vital that AI practitioners take the effective (not maximum) GPU throughput into account
    when devising their training plan using the Chinchilla scaling law.

Below we explore how na\"ively applying the Chinchilla law can lead to inaccurate compute-optimal model/token sizes.
We then describe how \proposed can be utilize to derive a realistic compute-optimal model size within a given compute budget.
Consider the following three parameters that describe the Chinchilla law:
    the compute-optimal number of model parameters ($N$) and the number of tokens required to sufficiently train such model ($T$),
    given a fixed amount of compute budget ($C$).
For the purpose of our discussion, let us assume a scenario where we can utilize $420$ NVIDIA DGX A100 server nodes~\cite{dgxa100}
    (amounting to 3,360 A100 GPUs) for $30$ days. 

\textbf{Na\"ive estimation of Chinchilla point.} According to the Chinchilla scaling law~\cite{chinchilla}, $N$ and $T$ are calculated based on the following equations. 
\[N = \alpha \times C ^{0.5}, \ T = \beta \times C^{0.5} \ (\alpha = 0.089, \beta = 1.875)\]

Na\"ively assuming $100\%$ GPU utility leads to an (overprovisioned) compute budget of $C = 2.72e\text{+}24$ FLOPs over the period of 30 days.
With such compute budget $C$, one would lead to the conclusion that a compute-optimal model size of $N$ $=$ $145.61$ billion parameters
    can be trained with $T$~$=$~2,912 billion tokens within 30 days.
However, \proposed reveals that, even with the most optimal parallelization, the training system in aggregate achieves an average $35.56\%$ GPU utility.
This leads to a total of $85$ days of training time to train $T$ tokens over the LLM with $N$ parameters,
    more than $3\times$ the originally expected $30$ days of training time.

\begin{table}[t!]
    \vspace{0.7em}
    \centering
    \caption{
        Compute-optimal Chinchilla points under different compute budgets.
        The ``Optimal (\emph{t,d,p})'' column represents the optimal 3D parallelism plan under the given compute budget and LLM model architecture.
    }
    \label{tab:case2}
    \resizebox{\columnwidth}{!}{
    \begin{tabular}{|c|c|c|c|c|c|}
        \hline
        \boldmath$h$ &
        \boldmath$L$ &
        \textbf{\begin{tabular}[c]{@{}c@{}}Parameters\\ (billion)\end{tabular}} &
        \textbf{\begin{tabular}[c]{@{}c@{}}Tokens\\ (billion)\end{tabular}} &
        \textbf{\begin{tabular}[c]{@{}c@{}}Optimal\\ (t,d,p)\end{tabular}} &
        \textbf{\begin{tabular}[c]{@{}c@{}}Estimated \\ training time (days)\end{tabular}} \\ \hline
        \textbf{12,288} & \textbf{80} & \textbf{145.61} & \textbf{2,912} & \textbf{(8,42,10)} & \textbf{85} \\ \hline
        12,288          & 70          & 127.49          & 2,550          & (8,60,7)           & 64          \\ \hline
        12,288          & 60          & 109.37          & 2,187          & (8,70,6)           & 47          \\ \hline
        10,240          & 70          & 88.62           & 1,772          & (8,84,5)           & 40          \\ \hline
        \textbf{10,240} & \textbf{60} & \textbf{76.04}  & \textbf{1,521} & \textbf{(8,70,6)} & \textbf{30} \\ \hline
        9,216           & 80          & 82.03           & 1,641          & (8,84,5)           & 37          \\ \hline
        9,216           & 70          & 71.83           & 1,437          & (8,84,5)           & 29          \\ \hline
        \end{tabular}%
    }
    \vspace{-0.8em}
\end{table}

\textbf{Realistically estimating compute-optimal Chinchilla point.}
We now discuss how \proposed can be utilized to evaluate a more realistic value of $N$ and $T$ by taking \emph{effective} GPU utilization into consideration. 
We begin by exploring different model sizes, varying the number of decoder layers ($L$) and hidden dimensions ($h$) of LLMs.
For a given LLM model configuration, we then exhaustively evaluate all possible parallelism strategies within the ($t$, $d$, $p$)-way 3D parallelism search space. 
By identifying the parallelism strategy that results in the smallest single-iteration training time,
    we check whether the end-to-end training time to consume the entire $T$ tokens falls within the ``effective'' compute budget $C$.
Among all candidate $N$, $T$, and $C$ Chinchilla points, the design point with the largest model size $N$
    becomes the most compute-optimal LLM with the highest algorithmic performance given the compute budget $C$.
\tab{tab:case2} summarizes such exploration results where the first row with the estimated training time of $85$ days is the Chinchilla point derived
    assuming a na\"ive $100\%$ GPU compute utility. 
Using \proposed, we arrive at a more realistic Chinchilla point having a model size of $76.04$ billion parameters trained with $1,521$ billion tokens within $30$ days,
    which is a $48\%$ smaller LLM model architecture compared to the na\"ively estimated version.

\section{Related Work}
\label{sect:related_work}

\textbf{Distributed training.} Several
studies~\cite{prague,jia2019,optimus_cc,dapple,zero,zero_off,zero_inf} explored
distributed training across multiple workers.  GPipe~\cite{gpipe} and
PipeDream~\cite{pipedream,pipedream_bw} propose layer-wise partitioning of
model parameters and scheduling approaches to effectively execute partitioned
layers in a pipelined manner. Megatron-LM~\cite{megatron_lm} proposed tensor
parallelism which is tailored for transformer-based language models.  There are
also prior works proposing advanced collective communication algorithms for
distributed training. FlexReduce~\cite{flexreduce},
BlueConnect~\cite{blueconnect}, and Themis~\cite{themis} proposed
novel bandwidth-aware All-Reduce algorithms on
heterogeneous/asymmetric network topology.
Putting these efforts together, Microsoft/NVIDIA discussed their distributed training
for Megatron-Turing NLG 530B~\cite{mt_nlg} and the 1T-parameter
GPT model~\cite{megatron2} on thousands of GPUs via 3D parallelism.

\textbf{Performance modeling for distributed training.}
\label{sect:related_work:performance_modeling}
While not targeting LLMs, ASTRA-sim~\cite{astra-sim} presents a cycle-level simulator targeting distributed training, focusing on
modeling collective communication algorithms involved in different multi-dimensional network topologies.
Instead of using the cycle-level network backend (i.e., Garnet~\cite{garnet}), ASTRA-sim2.0~\cite{astra-sim-2} employs an analytical model for network backend,
reducing simulation time.
There are also several studies focusing on accurately modeling performance for modern AI models,
    but these works often limit their scope by either designing analytical models~\cite{amped,two_cs,alpa,calculon}, 
    profiling a representative subset of operators or specific configurations to project the performance of others~\cite{two_cs,seqpoint,alpa},
    or targeting small scale training systems with only tens of GPUs~\cite{tofu,alpa,seqpoint}.
Specifically, AMPeD~\cite{amped} is an analytical model for estimating performance in distributed training of transformer-based models,
    but its performance modeling is controlled by a factor for compute core efficiency, which is driven by fitting empirical experimental results
    on various transformer-based models and underlying hardwares, sacrificing specificity for individual scenarios.
Conversely, Tale of Two Cs~\cite{two_cs} employs a focused methodology to understand computation and communication scaling in future transformers, 
    selecting representative operators for detailed profiling and using these to model performance of the entire training process across various model hyperparameters.
Calculon~\cite{calculon} proposes a parameterized analytical performance 
    model exploring optimal parallel training configurations. This work parameterizes the target LLM, system, and training configuration to produce results broken down into compute time and memory usage. Although validation was performed on a real multi-node GPU system, the validation was conducted over 8 data points only, and the outcomes from the suggested optimal search strategy have not been validated over real GPU systems. In contrast, \proposed and its practicality have been validated over various multi-node training systems as shown in \sect{sec-4} and \sect{sect:cost-optimal-case-study} (\tab{tab:mtnlg_small_scale}). More critically, Calculon's analytical model does not consider framework-level/low-level implementations of the target LLM's execution because it is based on analytical modeling of a very specific implementation of LLM execution engines (e.g., Megatron), unlike vTrain where any changes or software updates to the framework-level/low-level implementations of LLMs (e.g., an upgrade from FlashAttention~\cite{flashattn} to FlashAttention-2~\cite{flashattn2} to accelerate attention layers) will naturally be captured by vTrain as our work employs profiling for performance estimation.
\begin{table}[t!]
    \centering
    \vspace{0.7em}
    \caption{Comparison of \proposed vs. other performance models.}
    \label{tab:comparison}
    \resizebox{\columnwidth}{!}{%
    \begin{tabular}{|c|c|c|c|c|c|c|}
\hline
    &  
    \begin{tabular}[c]{@{}c@{}}\textbf{ASTRA-}\\\textbf{sim}~\cite{astra-sim}\end{tabular} & 
    \begin{tabular}[c]{@{}c@{}}\textbf{AMPeD}\\ \cite{amped}\end{tabular} & 
    \begin{tabular}[c]{@{}c@{}}\textbf{SeqPoint}\\ \cite{seqpoint}\end{tabular} & 
    \begin{tabular}[c]{@{}c@{}}\textbf{Tale of} \\\textbf{Two Cs}~\cite{two_cs}\end{tabular} &
    \begin{tabular}[c]{@{}c@{}}\textbf{Calculon}\\ \cite{calculon}\end{tabular} &
    \textbf{vTrain} \\ \hline
\textbf{Target workload} & Any & Transformer & RNN & Transformer & Transformer & Transformer \\ \hline
\begin{tabular}[c]{@{}c@{}}\textbf{Simulation time}\\\textbf{for a single}\\ \textbf{training iteration}\end{tabular} &  
    N/A & 
    seconds & 
    N/A & 
    N/A &
    milliseconds &
    seconds \\ \hline
\begin{tabular}[c]{@{}c@{}}\textbf{Performance} \\ \textbf{modeling}\end{tabular} &  
    \begin{tabular}[c]{@{}c@{}}cycle-\\ level\\ (analytical\\ for 2.0)\end{tabular} & 
    analytical & 
    \begin{tabular}[c]{@{}c@{}}profile-\\ based \\ (sampled)\end{tabular} & 
    \begin{tabular}[c]{@{}c@{}}profile-\\ based \\ (sampled)\end{tabular} &
    analytical &
    \begin{tabular}[c]{@{}c@{}}profile-\\ based \\ (entire)\end{tabular}  \\ \hline
\begin{tabular}[c]{@{}c@{}}\textbf{Applicable to}\\ \textbf{different models}\end{tabular} & 
    \cellcolor[HTML]{D3F5C7}O & 
    \cellcolor[HTML]{FFCCC9}X & 
    \cellcolor[HTML]{FFCCC9}X & 
    \cellcolor[HTML]{FFCCC9}X & 
    \cellcolor[HTML]{FFCCC9}X & 
    \cellcolor[HTML]{D3F5C7}O \\ \hline
\begin{tabular}[c]{@{}c@{}}\textbf{Multi-GPU} \\ \textbf{support}\end{tabular} & 
    \cellcolor[HTML]{D3F5C7}O & 
    \cellcolor[HTML]{D3F5C7}O &
    \cellcolor[HTML]{FFCCC9}X & 
    \cellcolor[HTML]{D3F5C7}O &
    \cellcolor[HTML]{D3F5C7}O & 
    \cellcolor[HTML]{D3F5C7}O \\ \hline
\begin{tabular}[c]{@{}c@{}}\textbf{Validated over}\\ \textbf{hundreds of GPUs}\end{tabular} & 
    \cellcolor[HTML]{FFCCC9}X & 
    \cellcolor[HTML]{D3F5C7}O & 
    \cellcolor[HTML]{FFCCC9}X & 
    \cellcolor[HTML]{FFCCC9}X & 
    \cellcolor[HTML]{D3F5C7}O & 
    \cellcolor[HTML]{D3F5C7}O \\ \hline
\begin{tabular}[c]{@{}c@{}}\textbf{Number of}\\ \textbf{validation points}\end{tabular} & 
    0 & 
    \begin{tabular}[c]{@{}c@{}} 12 (single)\\ 9 (multi)\end{tabular} & 
    18 & 
    0 & 
    8 (multi) & 
    \begin{tabular}[c]{@{}c@{}} 1,440 (single) \\ 112 (multi) \end{tabular} \\ \hline
\begin{tabular}[c]{@{}c@{}}\textbf{Average}\\ \textbf{validation error}\end{tabular} &  
    N/A & 
    $\sim$12\%  & 
    1.50\% & 
    N/A &
    3.65\% &
    \begin{tabular}[c]{@{}c@{}}8.37\% (single)\\ 14.73\% (multi)\end{tabular}\\ \hline
\end{tabular}%
}
\vspace{-0.8em}
\end{table}
SeqPoint~\cite{seqpoint} proposes performance modeling for training of sequence-based DNNs,
    selecting and profiling representative set of training iterations for the entire training process
    based on the key observation that hardware behavior differs according to the input sequence length.
While these works estimate the costs of un-profiled operators and configurations based on the profiling results of representative groups,
    \proposed basically profiles \emph{all operators}, including even short-living element-wise operations,
    but minimizes the profiling overhead by identifying repetitive operators that are inherently present in transformer-based LLM training algorithms.
In short, \proposed is specifically designed to evaluate modern LLMs requiring several thousands of GPUs for training,
    providing a fast yet highly accurate simulation toolchain for AI practitioners, one which we demonstrate using our case studies.
 \tab{tab:comparison} highlights key differences between \proposed and other performance models for AI systems.

\textbf{Exploiting deterministic dataflow of ML.} There has been prior
work employing a profiling-based approach to exploit the deterministic
execution property of modern AI algorithms for system optimizations.
Rammer~\cite{rammer} proposes a DNN compiler design that leverages
deterministic characteristics of operators in ML models to generate an
efficient operator-level schedule at compile time, maximizing hardware
utilization.  DeepPool~\cite{deeppool} employs a profiling based scheduling
to dynamically adjust GPU resource allocations for multi-tenant
training jobs.  PREMA~\cite{prema} and LazyBatching~\cite{lazybatching} utilize
the deterministic latency of AI inference to better schedule and batch input
queries in inference servers.  Daydream~\cite{daydream} proposes a simulation
methodology that estimates the efficacy of a diverse set DNN optimizations
(e.g., CPU-GPU tensor offloading, mixed precision training, etc).  None of
these prior works target modern, scale-out LLM training systems nor provide an
accurate prediction mechanism to evaluate cost-effective, compute-optimal LLM
training plans.

\textbf{Multi-tenant GPU clusters.} There is a long list of prior work focusing on efficient scheduling algorithms for multi-tenant GPU training clusters, either in a server-centric style~\cite{jeon19,hwang2021,bian21,hu21,gandiva_fair,antman,zhao22,heteroaware_cluster,gandiva,tiresias,chronus} or serverless style~\cite{optimus,themis_nsdi,lucid,sia,antman}.
Pollux~\cite{pollux} proposes a severless and elastic scheduling system, presenting an accurate analytical model for system goodput and statistical efficiency.
ElasticFlow~\cite{elasticflow} is a state-of-the-art serverless training platform that pre-measures training throughputs as a function of the number of GPUs and utilize that information to elastically allocate GPU resources intelligently. In general, the key contribution of \proposed is orthogonal to these prior art.
\section{Conclusion}

\proposed is a high-performance profiling-driven simulation
framework that accurately estimates LLM's training time. \proposed targets AI
systems equipped with thousands of GPUs for training LLM models
containing billions of parameters trained with trillions of tokens. We
demonstrate \proposed's merits and its practicality by conducting several case
studies, e.g., evaluation of optimal training parallelization
plans that balances training time and its associated cost, efficient
multi-tenant GPU cluster scheduling algorithms targeting multiple LLM training
jobs, and determining a cost-optimal LLM model architecture given a fixed
compute budget.

\section*{Acknowledgment}
This work was partly supported by the National Research Foundation of Korea (NRF) grant funded by the Korea government (MSIT) (NRF-2021R1A2C2091753),
by Institute of Information \& Communications Technology Planning \& Evaluation(IITP) grant funded by the Korea government(MSIT) (No.RS-2024-00402898, Simulation-based High-speed/High-Accuracy Data Center Workload/System Analysis Platform),
by Institute of Information \& Communications Technology Planning \& Evaluation(IITP) grant funded by the Korea government(MSIT) (No.RS-2024-00395134, DPU-Centric Datacenter Architecture for Next-Generation AI Devices),
by Institute of Information \& Communications Technology Planning \& Evaluation(IITP) grant funded by the Korea government(MSIT) (No.RS-2024-00438851, (SW Starlab) High-performance Privacy-preserving Machine Learning System and System Software),
and by the Samsung Electronics.
We also appreciate the support from Samsung Advanced Institute of Technology, Samsung Electronics Co., Ltd.
Minsoo Rhu is the corresponding author.

\bibliographystyle{IEEEtranS}
\bibliography{refs}

% Generated by IEEEtranS.bst, version: 1.13 (2008/09/30)
\begin{thebibliography}{10}
\providecommand{\url}[1]{#1}
\csname url@samestyle\endcsname
\providecommand{\newblock}{\relax}
\providecommand{\bibinfo}[2]{#2}
\providecommand{\BIBentrySTDinterwordspacing}{\spaceskip=0pt\relax}
\providecommand{\BIBentryALTinterwordstretchfactor}{4}
\providecommand{\BIBentryALTinterwordspacing}{\spaceskip=\fontdimen2\font plus
\BIBentryALTinterwordstretchfactor\fontdimen3\font minus
  \fontdimen4\font\relax}
\providecommand{\BIBforeignlanguage}[2]{{%
\expandafter\ifx\csname l@#1\endcsname\relax
\typeout{** WARNING: IEEEtranS.bst: No hyphenation pattern has been}%
\typeout{** loaded for the language `#1'. Using the pattern for}%
\typeout{** the default language instead.}%
\else
\language=\csname l@#1\endcsname
\fi
#2}}
\providecommand{\BIBdecl}{\relax}
\BIBdecl

\bibitem{garnet}
N.~Agarwal, T.~Krishna, L.-S. Peh, and N.~K. Jha, ``{GARNET: A Detailed On-Chip
  Network Model Inside a Full-System Simulator},'' in \emph{Proceedings of the
  IEEE International Symposium on Performance Analysis of Systems and Software
  (ISPASS)}, 2009.

\bibitem{amazon_p4d}
{AWS}, ``{Amazon EC2 P4 Instances},''
  \url{https://aws.amazon.com/ko/ec2/instance-types/p4/}, 2023.

\bibitem{bian21}
Z.~Bian, S.~Li, W.~Wang, and Y.~You, ``{Online Evolutionary Batch Size
  Orchestration for Scheduling Deep Learning Workloads in GPU Clusters},'' in
  \emph{Proceedings of the International Conference on High Performance
  Computing, Networking, Storage and Analysis (SC)}, 2021.

\bibitem{gpt3}
T.~Brown, B.~Mann, N.~Ryder, M.~Subbiah, J.~D. Kaplan, P.~Dhariwal,
  A.~Neelakantan, P.~Shyam, G.~Sastry, A.~Askell, S.~Agarwal, A.~Herbert-Voss,
  G.~Krueger, T.~Henighan, R.~Child, A.~Ramesh, D.~Ziegler, J.~Wu, C.~Winter,
  C.~Hesse, M.~Chen, E.~Sigler, M.~Litwin, S.~Gray, B.~Chess, J.~Clark,
  C.~Berner, S.~McCandlish, A.~Radford, I.~Sutskever, and D.~Amodei,
  ``{Language Models are Few-Shot Learners},'' in \emph{Proceedings of the
  International Conference on Neural Information Processing Systems (NeurIPS)},
  2020.

\bibitem{gandiva_fair}
S.~Chaudhary, R.~Ramjee, M.~Sivathanu, N.~Kwatra, and S.~Viswanatha,
  ``{Balancing Efficiency and Fairness in Heterogeneous GPU Clusters for Deep
  Learning},'' in \emph{Proceedings of the EuroSys Conference}, 2020.

\bibitem{blueconnect}
M.~Cho, U.~Finkler, D.~Kung, and H.~Hunter, ``{BlueConnect: Decomposing
  All-Reduce for Deep Learning on Heterogeneous Network Hierarchy},''
  \emph{Proceedings of Machine Learning and Systems}, 2019.

\bibitem{lazybatching}
Y.~Choi, Y.~Kim, and M.~Rhu, ``{Lazy Batching: An SLA-Aware Batching System for
  Cloud Machine Learning Inference},'' in \emph{Proceedings of the
  International Symposium on High-Performance Computer Architecture (HPCA)},
  2021.

\bibitem{prema}
Y.~Choi and M.~Rhu, ``{PREMA: A Predictive Multi-Task Scheduling Algorithm for
  Preemptible Neural Processing Units},'' in \emph{Proceedings of the
  International Symposium on High-Performance Computer Architecture (HPCA)},
  2020.

\bibitem{palm}
A.~Chowdhery, S.~Narang, J.~Devlin, M.~Bosma, G.~Mishra, A.~Roberts, P.~Barham,
  H.~W. Chung, C.~Sutton, S.~Gehrmann, P.~Schuh, K.~Shi, S.~Tsvyashchenko,
  J.~Maynez, A.~Rao, P.~Barnes, Y.~Tay, N.~Shazeer, V.~Prabhakaran, E.~Reif,
  N.~Du, B.~Hutchinson, R.~Pope, J.~Bradbury, J.~Austin, M.~Isard, G.~Gur-Ari,
  P.~Yin, T.~Duke, A.~Levskaya, S.~Ghemawat, S.~Dev, H.~Michalewski, X.~Garcia,
  V.~Misra, K.~Robinson, L.~Fedus, D.~Zhou, D.~Ippolito, D.~Luan, H.~Lim,
  B.~Zoph, A.~Spiridonov, R.~Sepassi, D.~Dohan, S.~Agrawal, M.~Omernick, A.~M.
  Dai, T.~S. Pillai, M.~Pellat, A.~Lewkowycz, E.~Moreira, R.~Child, O.~Polozov,
  K.~Lee, Z.~Zhou, X.~Wang, B.~Saeta, M.~Diaz, O.~Firat, M.~Catasta, J.~Wei,
  K.~Meier-Hellstern, D.~Eck, J.~Dean, S.~Petrov, and N.~Fiedel, ``{PaLM:
  Scaling Language Modeling with Pathways},'' \emph{arXiv preprint
  arXiv:2204.02311}, 2022.

\bibitem{flashattn2}
T.~Dao, ``{FlashAttention-2: Faster Attention with Better Parallelism and Work
  Partitioning},'' \emph{arXiv preprint arXiv:2307.08691}, 2023.

\bibitem{flashattn}
T.~Dao, D.~Fu, S.~Ermon, A.~Rudra, and C.~R\'{e}, ``{FlashAttention: Fast and
  Memory-Efficient Exact Attention with IO-Awareness},'' in \emph{Proceedings
  of the International Conference on Neural Information Processing Systems
  (NeurIPS)}, 2022.

\bibitem{dapple}
S.~Fan, Y.~Rong, C.~Meng, Z.~Cao, S.~Wang, Z.~Zheng, C.~Wu, G.~Long, J.~Yang,
  L.~Xia, L.~Diao, X.~Liu, and W.~Lin, ``{DAPPLE: A Pipelined Data Parallel
  Approach for Training Large Models},'' in \emph{Proceedings of the 26th ACM
  SIGPLAN Symposium on Principles and Practice of Parallel Programming
  (PPoPP)}, 2021.

\bibitem{chronus}
W.~Gao, Z.~Ye, P.~Sun, Y.~Wen, and T.~Zhang, ``{Chronus: A Novel Deadline-Aware
  Scheduler for Deep Learning Training Jobs},'' in \emph{Proceedings of the ACM
  Symposium on Cloud Computing (SoCC)}, 2021.

\bibitem{github_copilot}
{GitHub}, ``{GitHub Copilot},'' \url{https://github.com/features/copilot},
  2023.

\bibitem{genai_studio}
{Google Cloud}, ``{Introduction to Generative AI Studio},''
  \url{https://cloud.google.com/vertex-ai/docs/generative-ai/learn/generative-ai-studio},
  2023.

\bibitem{elasticflow}
D.~Gu, Y.~Zhao, Y.~Zhong, Y.~Xiong, Z.~Han, P.~Cheng, F.~Yang, G.~Huang,
  X.~Jin, and X.~Liu, ``{ElasticFlow: An Elastic Serverless Training Platform
  for Distributed Deep Learning},'' in \emph{Proceedings of the International
  Conference on Architectural Support for Programming Languages and Operating
  Systems (ASPLOS)}, 2023.

\bibitem{tiresias}
J.~Gu, M.~Chowdhury, K.~G. Shin, Y.~Zhu, M.~Jeon, J.~Qian, H.~Liu, and C.~Guo,
  ``{Tiresias: A {GPU} Cluster Manager for Distributed Deep Learning},'' in
  \emph{Proceedings of the USENIX Symposium on Networked Systems Design and
  Implementation (NSDI)}, 2019.

\bibitem{chinchilla}
J.~Hoffmann, S.~Borgeaud, A.~Mensch, E.~Buchatskaya, T.~Cai, E.~Rutherford,
  D.~de~Las~Casas, L.~A. Hendricks, J.~Welbl, A.~Clark, T.~Hennigan, E.~Noland,
  K.~Millican, G.~van~den Driessche, B.~Damoc, A.~Guy, S.~Osindero,
  K.~Simonyan, E.~Elsen, J.~W. Rae, O.~Vinyals, and L.~Sifre, ``{Training
  Compute-Optimal Large Language Models},'' \emph{arXiv preprint
  arXiv:2203.15556}, 2022.

\bibitem{hu21}
Q.~Hu, P.~Sun, S.~Yan, Y.~Wen, and T.~Zhang, ``{Characterization and Prediction
  of Deep Learning Workloads in Large-Scale GPU Datacenters},'' in
  \emph{Proceedings of the International Conference on High Performance
  Computing, Networking, Storage and Analysis (SC)}, 2021.

\bibitem{lucid}
Q.~Hu, M.~Zhang, P.~Sun, Y.~Wen, and T.~Zhang, ``{Lucid: A Non-Intrusive,
  Scalable and Interpretable Scheduler for Deep Learning Training Jobs},'' in
  \emph{Proceedings of the International Conference on Architectural Support
  for Programming Languages and Operating Systems (ASPLOS)}, 2023.

\bibitem{gpipe}
Y.~Huang, Y.~Cheng, A.~Bapna, O.~Firat, D.~Chen, M.~Chen, H.~Lee, J.~Ngiam,
  Q.~V. Le, Y.~Wu, and z.~Chen, ``{GPipe: Efficient Training of Giant Neural
  Networks using Pipeline Parallelism},'' \emph{Proceedings of the
  International Conference on Neural Information Processing Systems (NeurIPS)},
  2019.

\bibitem{hwang2021}
C.~Hwang, T.~Kim, S.~Kim, J.~Shin, and K.~Park, ``{Elastic Resource Sharing for
  Distributed Deep Learning},'' in \emph{Proceedings of the USENIX Symposium on
  Networked Systems Design and Implementation (NSDI)}, 2021.

\bibitem{calculon}
M.~Isaev, N.~Mcdonald, L.~Dennison, and R.~Vuduc, ``{Calculon: a Methodology
  and Tool for High-Level Codesign of Systems and Large Language Models},'' in
  \emph{Proceedings of the International Conference on High Performance
  Computing, Networking, Storage and Analysis (SC)}, 2023.

\bibitem{jeon19}
M.~Jeon, S.~Venkataraman, A.~Phanishayee, J.~Qian, W.~Xiao, and F.~Yang,
  ``{Analysis of {Large-Scale} {Multi-Tenant} {GPU} Clusters for {DNN} Training
  Workloads},'' in \emph{Proceedings of the USENIX Conference on Usenix Annual
  Technical Conference}, 2019.

\bibitem{jia2019}
Z.~Jia, M.~Zaharia, and A.~Aiken, ``{Beyond Data and Model Parallelism for Deep
  Neural Networks.}'' in \emph{Proceedings of Machine Learning and Systems},
  2019.

\bibitem{flexreduce}
J.~Lee, I.~Hwang, S.~Shah, and M.~Cho, ``{FlexReduce: Flexible All-Reduce for
  Distributed Deep Learning on Asymmetric Network Topology},'' in
  \emph{Proceedings of ACM/IEEE Design Automation Conference (DAC)}, 2020.

\bibitem{gshard}
D.~Lepikhin, H.~Lee, Y.~Xu, D.~Chen, O.~Firat, Y.~Huang, M.~Krikun, N.~Shazeer,
  and Z.~Chen, ``{GShard: Scaling Giant Models with Conditional Computation and
  Automatic Sharding},'' \emph{arXiv preprint arXiv:2006.16668}, 2020.

\bibitem{li13pytorch}
S.~Li, Y.~Zhao, R.~Varma, O.~Salpekar, P.~Noordhuis, T.~Li, A.~Paszke,
  J.~Smith, B.~Vaughan, P.~Damania, and S.~Chintala, ``{PyTorch Distributed:
  Experiences on Accelerating Data Parallel Training},'' \emph{arXiv preprint
  arXiv:2006.15704}, 2020.

\bibitem{prague}
Q.~Luo, J.~He, Y.~Zhuo, and X.~Qian, ``{Prague: High-Performance
  Heterogeneity-Aware Asynchronous Decentralized Training},'' in
  \emph{Proceedings of the International Conference on Architectural Support
  for Programming Languages and Operating Systems (ASPLOS)}, 2020.

\bibitem{rammer}
L.~Ma, Z.~Xie, Z.~Yang, J.~Xue, Y.~Miao, W.~Cui, W.~Hu, F.~Yang, L.~Zhang, and
  L.~Zhou, ``{Rammer: Enabling Holistic Deep Learning Compiler Optimizations
  with {rTasks}},'' in \emph{{Proceedings of the USENIX Symposium on Operating
  Systems Design and Implementation (OSDI)}}, 2020.

\bibitem{themis_nsdi}
K.~Mahajan, A.~Balasubramanian, A.~Singhvi, S.~Venkataraman, A.~Akella,
  A.~Phanishayee, and S.~Chawla, ``{Themis: Fair and Efficient {GPU} Cluster
  Scheduling},'' in \emph{Proceedings of the USENIX Symposium on Networked
  Systems Design and Implementation (NSDI)}, 2020.

\bibitem{rsc}
{Meta AI}, ``{Introducing the Introducing the AI Research SuperCluster },''
  \url{https://ai.facebook.com/blog/ai-rsc/}, 2022.

\bibitem{openai_service}
{Microsoft}, ``{Azure OpenAI Service},''
  \url{https://azure.microsoft.com/ko-kr/products/ai-services/openai-service},
  2023.

\bibitem{ms_cluster_trace}
{Microsoft}, ``{ElasticFlow Traces},''
  \url{https://github.com/microsoft/elasticflow-traces}, 2023.

\bibitem{megatron_deepspeed}
{Microsoft}, ``{Megatron-DeepSpeed},''
  \url{https://github.com/microsoft/Megatron-DeepSpeed}, 2023.

\bibitem{amped}
D.~Moolchandani, J.~Kundu, F.~Ruelens, P.~Vrancx, T.~Evenblij, and
  M.~Perumkunnil, ``{AMPeD: An Analytical Model for Performance in Distributed
  Training of Transformers},'' in \emph{{Proceedings of the International
  Symposium on Performance Analysis of Systems Software (ISPASS)}}, 2023.

\bibitem{pipedream}
D.~Narayanan, A.~Harlap, A.~Phanishayee, V.~Seshadri, N.~R. Devanur, G.~R.
  Ganger, P.~B. Gibbons, and M.~Zaharia, ``{PipeDream: Generalized Pipeline
  Parallelism for DNN Training},'' in \emph{Proceedings of the ACM Symposium on
  Operating System Principles (SOSP)}, 2019.

\bibitem{pipedream_bw}
D.~Narayanan, A.~Phanishayee, K.~Shi, X.~Chen, and M.~Zaharia,
  ``{Memory-Efficient Pipeline-Parallel DNN Training},'' in \emph{{Proceedings
  of the International Conference on Machine Learning (ICML)}}, 2021.

\bibitem{heteroaware_cluster}
D.~Narayanan, K.~Santhanam, F.~Kazhamiaka, A.~Phanishayee, and M.~Zaharia,
  ``{{Heterogeneity-Aware} Cluster Scheduling Policies for Deep Learning
  Workloads},'' in \emph{{Proceedings of the USENIX Symposium on Operating
  Systems Design and Implementation (OSDI)}}, 2020.

\bibitem{megatron2}
D.~Narayanan, M.~Shoeybi, J.~Casper, P.~LeGresley, M.~Patwary, V.~Korthikanti,
  D.~Vainbrand, P.~Kashinkunti, J.~Bernauer, B.~Catanzaro, A.~Phanishayee, and
  M.~Zaharia, ``{Efficient Large-Scale Language Model Training on GPU Clusters
  Using Megatron-LM},'' in \emph{Proceedings of the International Conference on
  High Performance Computing, Networking, Storage and Analysis (SC)}, 2021.

\bibitem{azure_supercom}
{NVIDIA}, ``{NVIDIA Teams With Microsoft to Build Massive Cloud AI Computer},''
  \url{https://nvidianews.nvidia.com/news/nvidia-microsoft-accelerate-cloud-enterprise-ai},
  2022.

\bibitem{nccl}
{NVIDIA}, ``{NVIDIA Collective Communications Library (NCCL)},''
  \url{"https://developer.nvidia.com/nccl"}, 2023.

\bibitem{dgxa100}
{NVIDIA}, ``{NVIDIA DGX A100},''
  \url{https://www.nvidia.com/en-us/data-center/dgx-a100/}, 2023.

\bibitem{nvlink}
{NVIDIA}, ``{NVLink and NVSwitch},''
  \url{https://www.nvidia.com/en-us/data-center/nvlink/}, 2023.

\bibitem{nccl-test}
{NVIDIA}, ``{Performance reported by NCCL tests},''
  \url{https://github.com/NVIDIA/nccl-tests/blob/master/doc/PERFORMANCE.md},
  2023.

\bibitem{gpt3_training_time}
{NVIDIA Developer}, ``{Scaling Language Model Training to a Trillion Parameters
  Using Megatron},''
  \url{https://developer.nvidia.com/blog/scaling-language-model-training-to-a-trillion-parameters-using-megatron/},
  2021.

\bibitem{cupti}
{NVIDIA Developer}, ``{CUPTI},''
  \url{https://docs.nvidia.com/cuda/cupti/index.html}, 2023.

\bibitem{gpt4}
OpenAI, ``{GPT-4 Technical Report},'' \emph{arXiv preprint arXiv:2303.08774},
  2023.

\bibitem{openai_codex}
{OpenAI}, ``{OpenAI Codex},'' \url{https://openai.com/blog/openai-codex}, 2023.

\bibitem{chatgpt}
L.~Ouyang, J.~Wu, X.~Jiang, D.~Almeida, C.~Wainwright, P.~Mishkin, C.~Zhang,
  S.~Agarwal, K.~Slama, A.~Ray, J.~Schulman, J.~Hilton, F.~Kelton, L.~Miller,
  M.~Simens, A.~Askell, P.~Welinder, P.~F. Christiano, J.~Leike, and R.~Lowe,
  ``{Training Language Models to Follow Instructions with Human Feedback},''
  \emph{Proceedings of the International Conference on Neural Information
  Processing Systems (NeurIPS)}, 2022.

\bibitem{deeppool}
S.~J. Park, J.~Fried, S.~Kim, M.~Alizadeh, and A.~Belay, ``{Efficient Strong
  Scaling Through Burst Parallel Training},'' \emph{Proceedings of Machine
  Learning and Systems}, 2022.

\bibitem{allreduce}
P.~Patarasuk and X.~Yuan, ``{Bandwidth Optimal All-Reduce Algorithms for
  Clusters of Workstations},'' \emph{Journal of Parallel and Distributed
  Computing}, vol.~69, no.~2, pp. 117--124, 2009.

\bibitem{two_cs}
S.~Pati, S.~Aga, M.~Islam, N.~Jayasena, and M.~D. Sinclair, ``{Tale of Two Cs:
  Computation vs. Communication Scaling for Future Transformers on Future
  Hardware},'' in \emph{Proceedings of the International Symposium on Workload
  Characterization (IISWC)}, 2023.

\bibitem{seqpoint}
S.~Pati, S.~Aga, M.~D. Sinclair, and N.~Jayasena, ``{SeqPoint: Identifying
  Representative Iterations of Sequence-Based Neural Networks},'' in
  \emph{{Proceedings of the International Symposium on Performance Analysis of
  Systems Software (ISPASS)}}, 2020.

\bibitem{optimus}
Y.~Peng, Y.~Bao, Y.~Chen, C.~Wu, and C.~Guo, ``{Optimus: An Efficient Dynamic
  Resource Scheduler for Deep Learning Clusters},'' in \emph{Proceedings of the
  EuroSys Conference}, 2018.

\bibitem{pytorch}
{PyTorch}, ``{PyTorch Documentation},''
  \url{"https://pytorch.org/docs/versions.html"}, 2023.

\bibitem{pollux}
A.~Qiao, S.~K. Choe, S.~J. Subramanya, W.~Neiswanger, Q.~Ho, H.~Zhang, G.~R.
  Ganger, and E.~P. Xing, ``{Pollux: Co-adaptive Cluster Scheduling for
  Goodput-Optimized Deep Learning},'' in \emph{{Proceedings of the USENIX
  Symposium on Operating Systems Design and Implementation (OSDI)}}, 2021.

\bibitem{gpt}
A.~Radford, K.~Narasimhan, T.~Salimans, and I.~Sutskever, ``{Improving Language
  Understanding by Generative Pre-Training},'' 2018.

\bibitem{gpt2}
A.~Radford, J.~Wu, R.~Child, D.~Luan, D.~Amodei, and I.~Sutskever, ``{Language
  Models are Unsupervised Multitask Learners},'' \emph{OpenAI blog}, 2019.

\bibitem{zero}
S.~Rajbhandari, J.~Rasley, O.~Ruwase, and Y.~He, ``{ZeRO: Memory optimizations
  Toward Training Trillion Parameter Models},'' in \emph{Proceedings of the
  International Conference on High Performance Computing, Networking, Storage
  and Analysis (SC)}, 2020.

\bibitem{zero_inf}
S.~Rajbhandari, O.~Ruwase, J.~Rasley, S.~Smith, and Y.~He, ``{ZeRO-Infinity:
  Breaking the GPU Memory Wall for Extreme Scale Deep Learning},'' in
  \emph{Proceedings of the International Conference on High Performance
  Computing, Networking, Storage and Analysis (SC)}, 2021.

\bibitem{astra-sim}
S.~Rashidi, S.~Sridharan, S.~Srinivasan, and T.~Krishna, ``{ASTRA-sim: Enabling
  SW/HW Co-Design Exploration for Distributed DL Training Platforms},'' in
  \emph{Proceedings of the IEEE International Symposium on Performance Analysis
  of Systems and Software (ISPASS)}, 2020.

\bibitem{themis}
S.~Rashidi, W.~Won, S.~Srinivasan, S.~Sridharan, and T.~Krishna, ``{Themis: A
  Network Bandwidth-Aware Collective Scheduling Policy for Distributed Training
  of DL Models},'' in \emph{Proceedings of the International Symposium on
  Computer Architecture (ISCA)}, 2022.

\bibitem{deepspeed}
J.~Rasley, S.~Rajbhandari, O.~Ruwase, and Y.~He, ``{DeepSpeed: System
  Optimizations Enable Training Deep Learning Models with Over 100 Billion
  Parameters},'' in \emph{Proceedings of the ACM SIGKDD International
  Conference on Knowledge Discovery \& Data Mining}, 2020.

\bibitem{zero_off}
J.~Ren, S.~Rajbhandari, R.~Y. Aminabadi, O.~Ruwase, S.~Yang, M.~Zhang, D.~Li,
  and Y.~He, ``{{ZeRO-Offload}: Democratizing {Billion-Scale} Model
  Training},'' in \emph{Proceedings of the USENIX Conference on Usenix Annual
  Technical Conference}, 2021.

\bibitem{megatron_lm}
M.~Shoeybi, M.~Patwary, R.~Puri, P.~LeGresley, J.~Casper, and B.~Catanzaro,
  ``{Megatron-LM: Training Multi-Billion Parameter Language Models using Model
  Parallelism},'' \emph{arXiv preprint arXiv:1909.08053}, 2019.

\bibitem{mt_nlg}
S.~Smith, M.~Patwary, B.~Norick, P.~LeGresley, S.~Rajbhandari, J.~Casper,
  Z.~Liu, S.~Prabhumoye, G.~Zerveas, V.~Korthikanti, E.~Zhang, R.~Child, R.~Y.
  Aminabadi, J.~Bernauer, X.~Song, M.~Shoeybi, Y.~He, M.~Houston, S.~Tiwary,
  and B.~Catanzaro, ``{Using DeepSpeed and Megatron to Train Megatron-Turing
  NLG 530B, a Large-Scale Generative Language Model},'' \emph{arXiv preprint
  arXiv:2201.11990}, 2022.

\bibitem{optimus_cc}
J.~Song, J.~Yim, J.~Jung, H.~Jang, H.-J. Kim, Y.~Kim, and J.~Lee,
  ``{Optimus-CC: Efficient Large NLP Model Training with 3D Parallelism Aware
  Communication Compression},'' in \emph{Proceedings of the International
  Conference on Architectural Support for Programming Languages and Operating
  Systems (ASPLOS)}, 2023.

\bibitem{sia}
S.~J. Subramanya, D.~Arfeen, S.~Lin, A.~Qiao, Z.~Jia, and G.~R. Ganger, ``{Sia:
  Heterogeneity-Aware, Goodput-Optimized ML-Cluster Scheduling},'' in
  \emph{Proceedings of the ACM Symposium on Operating System Principles
  (SOSP)}, 2023.

\bibitem{gemini}
G.~Team, R.~Anil, S.~Borgeaud, Y.~Wu, J.-B. Alayrac, J.~Yu, R.~Soricut,
  J.~Schalkwyk, A.~M. Dai, A.~Hauth \emph{et~al.}, ``{Gemini: a family of
  highly capable multimodal models},'' \emph{arXiv preprint arXiv:2312.11805},
  2023.

\bibitem{llama}
H.~Touvron, T.~Lavril, G.~Izacard, X.~Martinet, M.-A. Lachaux, T.~Lacroix,
  B.~Rozière, N.~Goyal, E.~Hambro, F.~Azhar, A.~Rodriguez, A.~Joulin,
  E.~Grave, and G.~Lample, ``{LLaMA: Open and Efficient Foundation Language
  Models},'' \emph{arXiv preprint arXiv:2302.13971}, 2023.

\bibitem{llama2}
H.~Touvron, L.~Martin, K.~Stone, P.~Albert, A.~Almahairi, Y.~Babaei,
  N.~Bashlykov, S.~Batra, P.~Bhargava, S.~Bhosale, D.~Bikel, L.~Blecher, C.~C.
  Ferrer, M.~Chen, G.~Cucurull, D.~Esiobu, J.~Fernandes, J.~Fu, W.~Fu,
  B.~Fuller, C.~Gao, V.~Goswami, N.~Goyal, A.~Hartshorn, S.~Hosseini, R.~Hou,
  H.~Inan, M.~Kardas, V.~Kerkez, M.~Khabsa, I.~Kloumann, A.~Korenev, P.~S.
  Koura, M.-A. Lachaux, T.~Lavril, J.~Lee, D.~Liskovich, Y.~Lu, Y.~Mao,
  X.~Martinet, T.~Mihaylov, P.~Mishra, I.~Molybog, Y.~Nie, A.~Poulton,
  J.~Reizenstein, R.~Rungta, K.~Saladi, A.~Schelten, R.~Silva, E.~M. Smith,
  R.~Subramanian, X.~E. Tan, B.~Tang, R.~Taylor, A.~Williams, J.~X. Kuan,
  P.~Xu, Z.~Yan, I.~Zarov, Y.~Zhang, A.~Fan, M.~Kambadur, S.~Narang,
  A.~Rodriguez, R.~Stojnic, S.~Edunov, and T.~Scialom, ``{Llama 2: Open
  Foundation and Fine-Tuned Chat Models},'' \emph{arXiv preprint
  arXiv:2307.09288}, 2023.

\bibitem{transformer}
A.~Vaswani, N.~Shazeer, N.~Parmar, J.~Uszkoreit, L.~Jones, A.~N. Gomez,
  {\L}.~Kaiser, and I.~Polosukhin, ``{Attention is All You Need},''
  \emph{Proceedings of the International Conference on Neural Information
  Processing Systems (NeurIPS)}, 2017.

\bibitem{tofu}
M.~Wang, C.-c. Huang, and J.~Li, ``{Supporting Very Large Models using
  Automatic Dataflow Graph Partitioning},'' in \emph{Proceedings of the EuroSys
  Conference}, 2019.

\bibitem{astra-sim-2}
W.~Won, T.~Heo, S.~Rashidi, S.~Sridharan, S.~Srinivasan, and T.~Krishna,
  ``{ASTRA-sim2.0: Modeling Hierarchical Networks and Disaggregated Systems for
  Large-model Training at Scale},'' in \emph{{Proceedings of the International
  Symposium on Performance Analysis of Systems Software (ISPASS)}}, 2023.

\bibitem{gandiva}
W.~Xiao, R.~Bhardwaj, R.~Ramjee, M.~Sivathanu, N.~Kwatra, Z.~Han, P.~Patel,
  X.~Peng, H.~Zhao, Q.~Zhang, F.~Yang, and L.~Zhou, ``{Gandiva: Introspective
  Cluster Scheduling for Deep Learning},'' in \emph{{Proceedings of the USENIX
  Symposium on Operating Systems Design and Implementation (OSDI)}}, 2018.

\bibitem{antman}
W.~Xiao, S.~Ren, Y.~Li, Y.~Zhang, P.~Hou, Z.~Li, Y.~Feng, W.~Lin, and Y.~Jia,
  ``{{AntMan}: Dynamic Scaling on {GPU} Clusters for Deep Learning},'' in
  \emph{{Proceedings of the USENIX Symposium on Operating Systems Design and
  Implementation (OSDI)}}, 2020.

\bibitem{opt}
S.~Zhang, S.~Roller, N.~Goyal, M.~Artetxe, M.~Chen, S.~Chen, C.~Dewan, M.~Diab,
  X.~Li, X.~V. Lin, T.~Mihaylov, M.~Ott, S.~Shleifer, K.~Shuster, D.~Simig,
  P.~S. Koura, A.~Sridhar, T.~Wang, and L.~Zettlemoyer, ``{OPT: Open
  Pre-Trained Transformer Language Models},'' \emph{arXiv preprint
  arXiv:2205.01068}, 2022.

\bibitem{zhao22}
Y.~Zhao, Y.~Liu, Y.~Peng, Y.~Zhu, X.~Liu, and X.~Jin, ``{Multi-Resource
  Interleaving for Deep Learning Training},'' in \emph{Proceedings of the ACM
  SIGCOMM Conference}, 2022.

\bibitem{alpa}
L.~Zheng, Z.~Li, H.~Zhang, Y.~Zhuang, Z.~Chen, Y.~Huang, Y.~Wang, Y.~Xu,
  D.~Zhuo, E.~P. Xing, J.~E. Gonzalez, and I.~Stoica, ``{Alpa: Automating
  Inter-and Intra-Operator Parallelism for Distributed Deep Learning},'' in
  \emph{{Proceedings of the USENIX Symposium on Operating Systems Design and
  Implementation (OSDI)}}, 2022.

\bibitem{daydream}
H.~Zhu, A.~Phanishayee, and G.~Pekhimenko, ``{Daydream: Accurately Estimating
  the Efficacy of Optimizations for DNN Training},'' in \emph{Proceedings of
  the USENIX Conference on Usenix Annual Technical Conference}, 2020.

\end{thebibliography}

\end{document}